\documentclass[10pt,journal,compsoc]{IEEEtran}
\usepackage{color}
\usepackage[utf8]{inputenc}
\usepackage{graphicx}
\usepackage[table,xcdraw]{xcolor}
\usepackage{multirow}
\usepackage{amsmath}
\usepackage{amssymb}
\usepackage{amsthm}
\usepackage{mathrsfs}
\usepackage{bm}
\usepackage[noend]{algpseudocode}
\usepackage[ruled,linesnumbered]{algorithm2e}
\usepackage{subfigure}
\usepackage{setspace}
\renewcommand{\thetable}{\arabic{table}}

\ifCLASSOPTIONcompsoc
  \usepackage[nocompress]{cite}
\else
  \usepackage{cite}
  
\fi

\ifCLASSINFOpdf
\else
\fi
\begin{document}

\title{Query-Efficient Adversarial Attack Against Vertical Federated Graph Learning}

\author{Jinyin Chen,
        Wenbo Mu,
        Luxin Zhang,
        Guohan Huang,
        Haibin Zheng,
        Yao Cheng
     
\IEEEcompsocitemizethanks{\IEEEcompsocthanksitem Jinyin Chen is with the Institute of Cyberspace Security, the College of Information
Engineering, Zhejiang University of Technology, Hangzhou, 310023, China. (e-mail: chenjinyin@zjut.edu.cn).
\IEEEcompsocthanksitem Wenbo Mu is with the College of Information Engineering, Zhejiang University of Technology, Hangzhou 310023, China. (e-mail: 211123030043@zjut.edu.cn).
\IEEEcompsocthanksitem Luxin Zhang is currently an Assistant Engineer with National Key Laboratory of Electromagnetic Space Security, Jiaxing, China. (e-mail: lxzhangMr@126.com).
\IEEEcompsocthanksitem Guohan Huang is with the College of Information Engineering, Zhejiang University of Technology, Hangzhou 310023, China. (e-mail: hgh0545@163.com).
\IEEEcompsocthanksitem Haibin Zheng is with the College of Computer Science and Technology, the Institute of Cyberspace Security, Zhejiang University of Technology, Hangzhou, 310023, China. (e-mail: haibinzheng320@gmail.com)
\IEEEcompsocthanksitem Yao Cheng is with the Digital Service,T{\"U}V S{\"U}D Asia Pacific Pte. Ltd., 609937, Singapore (e-mail: yao.cheng@tuvsud.com).
}

\thanks{Corresponding author: Haibin Zheng.}

}

\markboth{IEEE TRANSACTIONS ON NETWORK SCIENCE AND ENGINEERING}%
{Shell \MakeLowercase{\textit{et al.}}: Bare Demo of IEEEtran.cls for Computer Society Journals}

\IEEEtitleabstractindextext{
\begin{abstract}

Graph neural network (GNN) has captured wide attention due to its capability of graph representation learning for graph-structured data. However, the distributed data silos limit the performance of GNN. Vertical federated learning (VFL), an emerging technique to process distributed data, successfully makes GNN possible to handle the distributed graph-structured data. Despite the prosperous development of vertical federated graph learning (VFGL), the robustness of VFGL against the adversarial attack has not been explored yet. Although numerous adversarial attacks against centralized GNNs are proposed, their attack performance is challenged in the VFGL scenario. To the best of our knowledge, this is the first work to explore the adversarial attack against VFGL. A query-efficient hybrid adversarial attack framework is proposed to significantly improve the centralized adversarial attacks against VFGL, denoted as \emph{NA$^{2}$}, short for \emph{\underline{N}euron-based \underline{A}dversarial \underline{A}ttack}. Specifically, a malicious client manipulates its local training data to improve its contribution in a stealthy fashion. Then a shadow model is established based on the manipulated data to simulate the behavior of the server model 
in VFGL. As a result, the shadow model can improve the attack success rate of various centralized attacks with a few queries. Extensive experiments on five real-world benchmarks demonstrate that NA$^{2}$ improves the performance of the centralized adversarial attacks against VFGL, achieving state-of-the-art performance even under potential adaptive defense where the defender knows the attack method. Additionally, we provide interpretable experiments of the effectiveness of NA$^{2}$ via sensitive neurons identification and visualization of t-SNE. The code and datasets are available at https://github.com/hgh0545/NA2.
\end{abstract}

}

\maketitle

\IEEEdisplaynontitleabstractindextext

%
\IEEEpeerreviewmaketitle

\IEEEraisesectionheading{\section{Introduction}\label{sec:introduction}}

%
%

%
%

\IEEEPARstart{F}{ederated} graph learning (FGL), a novel distributed learning paradigm, establishes graph neural networks (GNNs) based on distributed data while keeping it decentralized, for the privacy-preserving concern. FGL can be categorized into horizontal FGL (HFGL) and vertical FGL (VFGL) in line with the data distribution characteristic. HFGL~\cite{zheng2021asfgnn, wang2020graphfl} is designed for the scenario where clients share the same feature space but different node ID space. Under the vertical setting, VFGL~\cite{zhou2020vertically, ni2021vertical} is suitable for the scenario where clients share the same node ID space but different feature space.

\begin{figure}[t]
  \centering 
  \includegraphics[width=3.5 in]{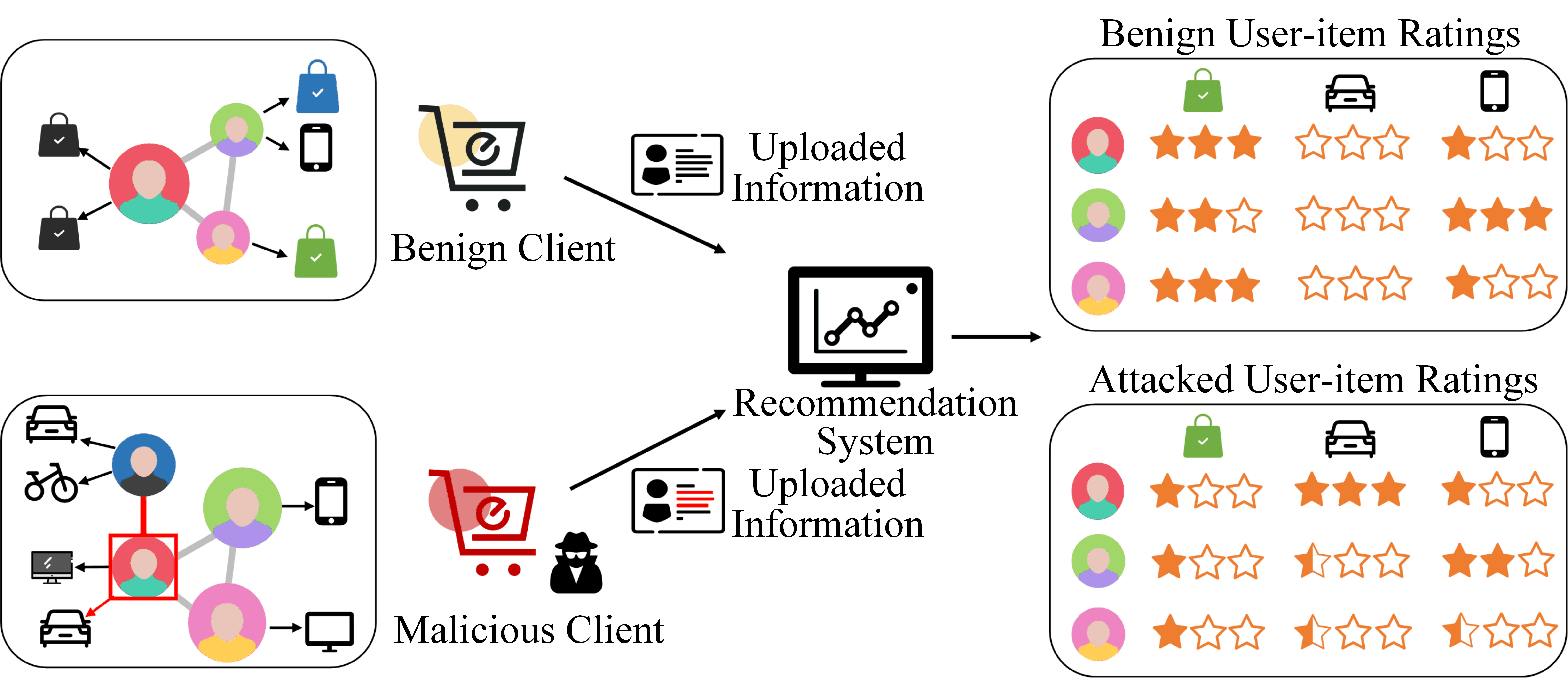}
  \caption{An illustration of the adversarial attack against VFGL in the e-commerce scenario. A small perturbation (e.g., fake friendships, browsing records) may mislead the analysis model and cause the wrong recommendation.}
  \label{fig:intro} 
\end{figure}
\setlength{\belowcaptionskip}{-1cm}

\begin{figure*}
  \centering 
  \includegraphics[width=5.1 in]{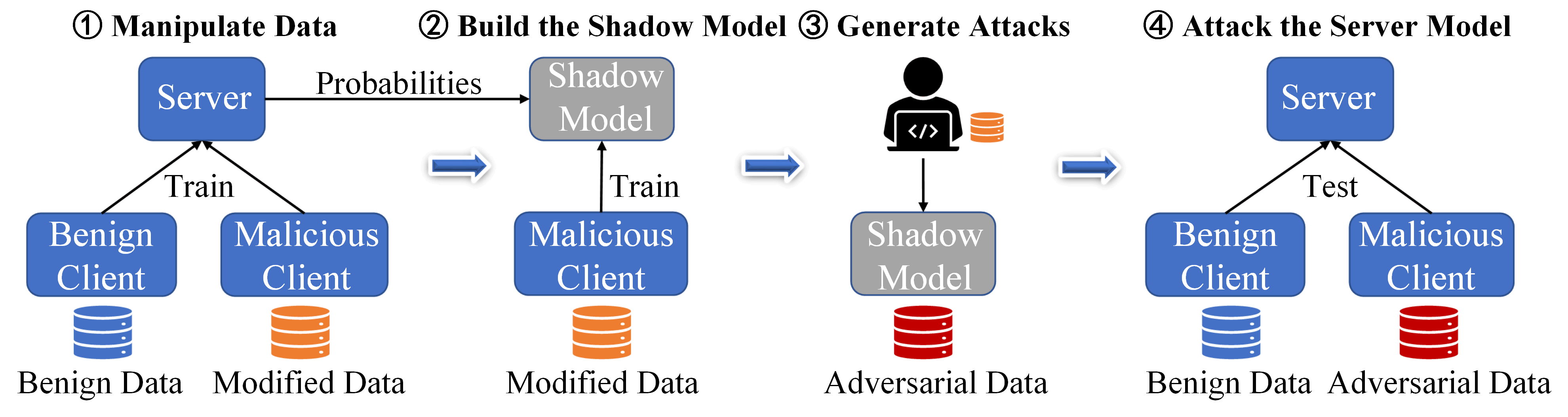}
  \caption{The pipeline of query-efficient hybrid adversarial attack framework against VFGL.}
  \label{fig:overview} 
\end{figure*}
\setlength{\belowcaptionskip}{-1cm}

Vertical data distribution is common in practical applications. For example, in electronic commerce, different e-commerce platforms may have the same users, but different attributes space due to the small intersection of the user behaviors over different products and services, such as the consumption behavior or browsing records. It is beneficial for the e-commerce platforms to train a reliable model (i.e., the server model) collaboratively to analyze the users' behavior and to achieve a win-win situation. VFGL is suitable for such a practical scenario, where the data can be constructed into graphs without sharing sensitive data. 

In such a practical application, there is still a potential threat in VFGL, i.e., it may be exposed to the threat of adversarial attacks. Fig.~\ref{fig:intro} shows the threat model in a practical scenario. Some dishonest e-commerce platforms may manipulate the data for the purpose of unfair competition. For example, adversarial attacks cause other cooperative clients to obtain incorrect predictive feedback when accessing the server model (e.g., modeling the user interest) in VFGL. As a result, erroneous predictions of the users' interests may degrade the counterparty’s recommendation system.

\textcolor{black}{In this work, we focus on exploring the vulnerability of VFGL to adversarial attacks.} Numerous adversarial attacks have been proposed against the centralized GNN (i.e., the GNN with centralized data), including white-box attacks~\cite{chen2018fast,li2021adversarial, wu2019adversarial}, gray-box attacks~\cite{zugner2018adversarial, zugner2019adversarial} and black-box attacks~\cite{dai2018adversarial, chang2022adversarial, ma2020towards, ma2021graph}. 
Although these attacks conduct successful attacks on the centralized GNN, they are all challenged in VFGL scenarios. \textcolor{black}{The white-box and gray-box attacks are not applicable to VFGL,} since they require the knowledge of the target model's structure, parameters, or both. However, the malicious client in VFGL does not have access to the server model which makes the final decision.  
For black-box attacks, one of the mainstream methods is  constructing a shadow model, use the shadow model to generate adversarial examples,  then transfer them to the target model. Another type is implemented by optimizing adversarial examples evaluated by the feedback of the target model. The latter relies on a large number of queries to the server model, which incurs high query costs and increases the risk of being detected. Thus, we select to build the shadow model as the strategy for adversarial attacks against VFGL.

In summary, there are some challenges for adversarial attacks against VFGL. (i) \emph{Knowledge Limitation}: It requires constructing successful attacks without any structure or parameter details of the server model.
(ii) \emph{Query Limitation}: The queries for prediction results from the server should be limited for stealthiness.
To address these challenges, we propose an adversarial attack framework against VFGL through a four-stage pipeline, shown in Fig.~\ref{fig:overview}. To tackle challenge (i), we construct a reliable shadow model during the training process. \textcolor{black}{The training data of the local client remains private to the server and other clients}, the malicious client can manipulate it to make the server model more dependent on the malicious client's local data, which is beneficial to use the malicious client's local data to train a shadow model similar to the server model. Thus, the adversary can use the shadow model to generate adversarial examples without query from the server, to address challenge (ii).

The contributions of this work are summarized as follows:

\begin{itemize}
\item To the best of our knowledge, this is the first work of adversarial attack against VFGL. To address the ineffectiveness of centralized attacks against VFGL, we proposed a query-efficient adversarial attack framework, namely NA$^{2}$, which can facilitate a significant improvement for centralized adversarial attacks.
\item Inspired by the observation that the server model in VFGL is more likely to be misled by high-contribution clients, a training data manipulation strategy is adopted for NA$^{2}$ to enhance the adversarial attack. Thus, NA$^{2}$ is proposed as a novel hybrid attack, involving training data manipulation, and adversarial attack generation in the testing process.
\item By constructing a reliable shadow model in the malicious client, only one query to the server is needed in NA$^{2}$, which makes NA$^{2}$ more efficient and effective.
\item Extensive experiments are conducted on five real-world graph-structured datasets, and four advanced adversarial attacks are evaluated in VFGL. The results testify that NA$^{2}$ achieves state-of-the-art (SOTA) performance. Besides, NA$^{2}$ still remains a threat to the VFGL with the potential defense mechanism. Furthermore, we interpret the effectiveness of NA$^{2}$ using sensitive neurons identification and visualization of t-SNE.
\end{itemize}

The rest of the paper is organized as follows. Related works on federated graph learning, adversarial attacks on GNNs and privacy leakage on graph learning are reviewed in Section~\ref{RW}. In Section~\ref{PPF}, we present the preliminary and problem formulation. Then, the details of NA$^{2}$ are shown in Section~\ref{method}, and NA$^{2}$ is evaluated in Section~\ref{experiments}. In Section~\ref{discussion}, we discuss the limitation of NA$^{2}$, future work and the comparison with the backdoor attacks. At last, the paper is concluded in Section~\ref{conclusion}.

\section{Related Work}
Our work focuses on adversarial attacks against privacy-preserving VFGL. In this section, we begin with the existing works of the VFGL, and then the adversarial attacks against GNNs and privacy leakage on graph learning are summarized.
\label{RW}

\subsection{Adversarial Attacks against GNNs}
Numerous studies have shown that GNNs are vulnerable to adversarial attacks, which are well-designed and imperceptible. Based on the attacker's knowledge and capabilities, the existing works can be divided into white-box attacks, gray-box attacks and black-box attacks.

\textbf{White-Box Attacks:} In white-box settings, the attackers own the parameters of the target model, training data and the ground truth, etc. \textcolor{black}{The gradient-based attack method is a typical white-box attack}~\cite{chen2018fast, wu2019adversarial,  zang2020graph, li2021adversarial, chen2022graphattck}, which applies the gradient information of the target model to generate the adversarial perturbations. 

\textbf{Gray-Box Attacks:} For the gray-box settings, the attackers try to train a shadow model to approximate the target model. For instance, Zügner et al.~\cite{zugner2018adversarial} trained a simplified graph convolutional network for Nettack, which computes the misclassification loss sequentially for each edge in the candidate set. Besides, Metattack~\cite{zugner2019adversarial} is proposed based on the meta-gradient of the shadow model. 

\textbf{Black-Box Attacks:} In terms of the black-box attacks, reinforcement learning is a typical and widely-used technology. Dai et al.~\cite{dai2018adversarial} proposed RL-S2V, which applies the reinforcement learning and only requires the prediction labels from the target model. Also, genetic algorithm is applied in the black-box settings where the prediction is available~\cite{dai2018adversarial}. Ma et al.~\cite{ma2021graph} proposed a reinforcement learning based attack method named RaWatt, which preserves the properties of the graph. Chang et al.~\cite{chang2022adversarial} proposed GF-Attack, which performs the attack on the graph filter without accessing any knowledge of the target model. Ma et al.~\cite{ma2020towards} extended the common gradient-based attack to black-box settings via the relationship between gradient and PageRank.

The above-mentioned adversarial attack methods are mainly realized by modifying the edges/features of the nodes. Additionally, graph injection attacks are gaining attention. They focus on inject the malicious nodes into the graph rather than modifying the existing edges/features~\cite{sun2019node, zou2021tdgia, tao2021single}.

\subsection{Privacy Leakage on Graph Learning}
In this paper, the privacy leaked in probabilities \textcolor{black}{(posterior probabilities or confidence scores)} is exploited to construct the shadow model. Therefore, we summarize the privacy leakage on FGL. Unfortunately, the research on privacy leakage on FGL~\cite{chen2024dataleakage} is still in its fancy. Thus, the privacy leakage research on GNNs is mainly summarized, as well as existing privacy studies on FGL.

\textbf{Privacy Leakage on GNNs:} Recently, graph neural networks (GNNs)  have attracted tremendous attention  from both academia and industry~\cite{wang2027robustgnn,dong2024Ego-Aware} ,but GNNs are facing the serious challenge in privacy leakage. According to the goal, it is mainly divided into three main directions~\cite{zhang2021graphmi}: membership inference attack, model extraction attack and model inversion attack. For membership inference attack, Duddu et al.~\cite{duddu2020quantifying} proposed two membership attacks considering white-box and black-box settings. The white-box attack exploits the intermediate low dimensional embeddings, while the black-box one exploits the statistical difference in predictions of the model. He et al.~\cite{he2021node} showed that the target nodes with higher subgraph density are easier to perform membership inference, which is caused by the fact that dense subgraphs encourage the target nodes to participate more in the aggregation process of GNN training. Olatunji et al.~\cite{olatunji2021membership} considered additional structural information is the main reason for GNN privacy leakage. In terms of model extraction, Shen et al.~\cite{shen2021model} used the accuracy and fidelity to ensure that the shadow model can better imitate the behavior of the target model to achieve the privacy information of the target model. As for model inversion attack, He et al. ~\cite{he2021stealing} first proposed a graph edge stealing method based on the target GNN's output to infer whether there is an edge connection between any pair of nodes in the graph. GraphMI~\cite{zhang2021graphmi} is proposed to reconstruct the graph considering the properties of the graph including sparsity and feature smoothness. Besides, Zhang et al.~\cite{zhang2022inference} systematically studied the privacy leakage problem of graph embeddings, and successfully inferred the basic properties of the target graph, such as the number of nodes, edges, and graph density, etc. Some existing methods rely on obtaining the shadow datasets to build shadow models. However, it is difficult to obtain the shadow datasets with similar distributions to the original datasets in reality (e.g., dimension mismatch due to different class numbers, etc.).

\textbf{Privacy Leakage on FGL:} By sharing model parameters or node embeddings, FGL still suffers the threat of privacy leakage. Some works attempt to apply some privacy-preserving mechanisms to address this issue, such as secure multi-party computation~\cite{zhou2020vertically, chen2021fedgraph}, secret sharing~\cite{rizk2021graph, zheng2021asfgnn}, differential privacy~\cite{zhang2021fastgnn, wu2021fedgnn} and homomorphic encryption~\cite{ni2021vertical}, etc.

\subsection{Vertical Federated Graph Learning}
\textcolor{black}{Traditional GNNs, which rely on centralized data storage, face systemic privacy risks and high costs. Consequently, they struggle to meet the stringent privacy preservation requirements of modern applications. To address this issue, VFGL has emerged as a promising research area, particularly in vertical data distribution scenarios.}

 Zhou et al.~\cite{zhou2020vertically} proposed the first vertical learning paradigm to preserve the privacy, using the secure muti-party computation. Ni et al.~\cite{ni2021vertical} applied additively homomorphic encryption in the proposed framework named FedVGCN for node classification task. Besides, VFGL is widely used in practical tasks such as knowledge graph~\cite{chen2021fede} and financial fraud detection~\cite{suzumura2019towards}.

\textcolor{black}{\subsection{Attack and Defense in Vertical Federated  Learning}
Existing attacks on VFL have been shown to undermine model robustness~\cite{yang2020vertical}. For instance, some backdoor attacks can mislead the VFL model or damage its overall performance on the original task by designing malicious backdoors. Among them, Liu et al~\cite{liu2020backdoor} proposed a Label Replacement Backdoor attack (LRB) in which the attacker replaces the gradients of a triggered sample with the ones of a clean sample of the targeted class. Yuan et al~\cite{pang2022attacking} introduced the Adversarial Dominating Input (ADI), which is an input sample with features that override all other features and lead to certain model output. As for the adversarial attack against the VFGL, Graph-Fraudster~\cite{chen2022graph} is an attack method that generates adversarial perturbations on the local graph structure to mislead the server model and achieve incorrect predictions. For defense, RVFR~\cite{liu2021rvfr}, a novel framework for robust vertical federated learning, which defends against backdoor and adversarial feature attacks through feature subspace recovery and purification. CAE~\cite{zou2022defending} uses an autoencoder with entropy regularization to disguise true labels, preventing attackers from inferring labels based on batch-level gradients. To further improve robustness, the authors introduce DCAE, which combines CAE with DiscreteSGD to protect against label inference attacks. Unlike the independent feature vectors in VFL, the embeddings of the nodes in the graph are interrelated and interact with each other. Therefore the above defense methods cannot be directly applied to GVFL.}

\section{Threat Model}
\label{PPF}

Before presenting our method, we give the definition of VFGL, as well as the adversarial attacks on it. Then the threat model is introduced. For convenience, the used definitions of symbols are briefly listed in TABLE~\ref{Symbol}.
\begin{table}[!ht]
    \renewcommand\arraystretch{0.9}
    \centering
    \caption{The definitions of symbols.}
    \label{Symbol}
    \resizebox{9cm}{!}{
    \begin{tabular}{c|r}
    \hline \hline
    \textbf{Symbol}&\textbf{Definition}\\ \hline
    $G=(V,E)$ &the original graph with sets of nodes and edges \\
    $A,\hat{A}$ & the original / adversarial adjacency matrix of $G$\\
    $X,\hat{X},\Tilde{X}$ & the original / adversarial / manipulated feature matrix of nodes\\
    ${f}_{\theta }(\cdot)$ & the GNN model with parameter $\theta$\\
    $K$&the number of clients\\
    $\hat{h}_{m}$ & the node embeddings of the malicious client \\
    $h_{global},\hat{h}_{global}$ & the benign / adversarial global node embeddings\\
    $V_{L}$ & the set of nodes with labels\\
    $v_t$ & the target node to be attacked\\
    $y_{v_t}$ & the ground truth of the target node $v_t$\\
    $Y,Y^{'}$ & the label / probabilities list\\
    $|F|$ & the number of class for nodes in the $G$\\
    $\mathcal{S}(\cdot)$ & the server model in VFGL\\
    $\mathcal{\Tilde{S}}(\cdot)$ & the shadow model\\
    $I_N$ & the identity matrix\\
    $\Tilde{D}$ & the degree matrix\\
    $H^l$ & the $l$-th hidden representation of the local GNN\\
    $Z^l$ & the $l$-th output of the neurons\\
    $W^l$ & the $l$-th weight matrix of the local GNN\\
    $\tau$ & the starting epoch of NA$^2$\\
    $k^l$ & the index of selected neuron in the GNN's $l$-th layer\\
    $P_t$ & the neuron path of the target node $t$ in neuron testing\\
    $C$ & the set of the candidate nodes\\
    $I$ & the set of the target features\\
    $p$ & the probabilities returned by the server model\\
    $B_{\mathcal{V}}$ & the decision boundary of the classifier $\mathcal{V}$\\
    $\Delta$ & the attack budget for edges modification\\
    $\gamma$ & the scale for features modification\\
    $Q$ & the query budge for attackers\\
    \hline \hline
    \end{tabular}}
\end{table}

\begin{figure*}
  \centering 
  \includegraphics[width=5.3 in]{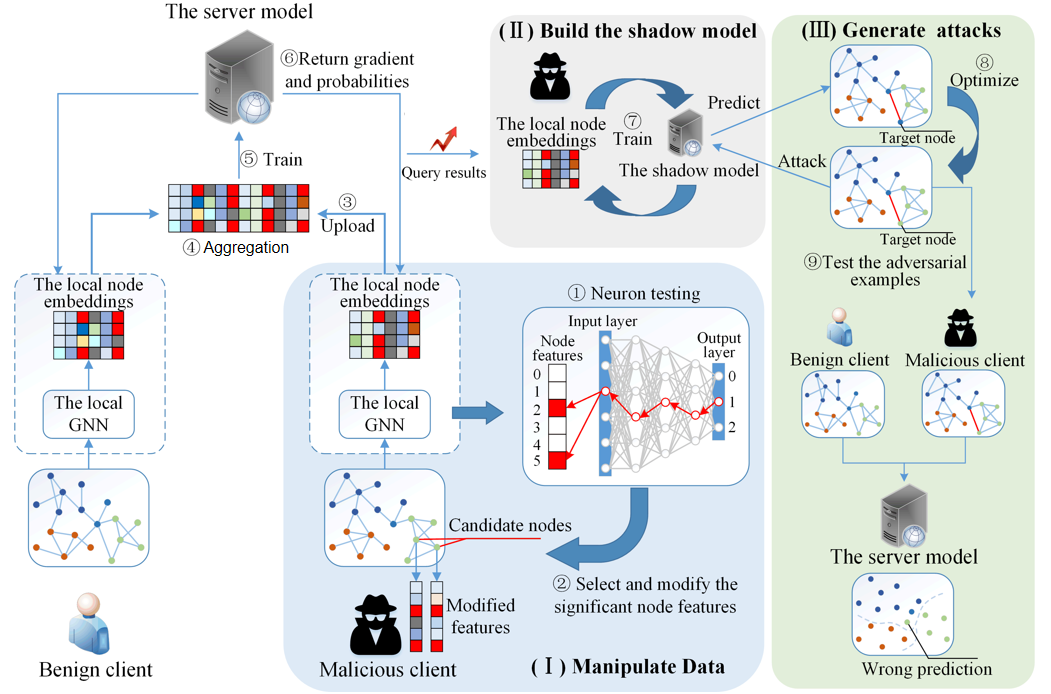}
  \textcolor{black}{\caption{Framework of NA$^{2}$. It can be divided into three stages: (i) Manipulate the local data according to neuron testing results. (ii) Build the shadow model by using the manipulated data. (iii) Generate the adversarial attacks via the shadow model. }}
  \label{fig:framwork} 
\end{figure*}

\textbf{Scenario:} In this work, a central server model is trained by multiple clients collaboratively in VFGL for node classification. In the training process, the server model issues the prediction and personalized gradient information to each client. During the testing process, the server model only provides the querying service and returns the probability. Besides, in the both training and testing process, the structure and inner parameters of the server model are not revealed to the clients.

\textbf{Adversary knowledge:} Assume that the malicious client only holds its own data (e.g., the adjacency, node features, and the ground truth label of the training nodes), and the information of the local GNN without collecting more extra information from the other clients.

\textbf{Attack goal:} \textcolor{black}{The malicious client aims to mislead the server model by uploading adversarial node embeddings, causing benign clients to receive incorrect predictions.} It should be noted that we only consider generating adversarial node embeddings through local GNN instead of constructing fake node embeddings directly because constructing fake node embeddings requires sufficient quality feedback from the server model, which requires multiple queries.
 Additionally, the preliminaries and problem formulation \textcolor{black}{for} Vertical Federated Learning and Adversarial Attacks on Vertical Federated Learning \textcolor{black}{are} presented in Appendix A.

\section{Methodology}
\label{method}

The malicious client can manipulate edges, features, or both. In this work, NA$^{2}$ \textcolor{black}{primarily targets the modification of training node features. This approach is preferred over modifying edges because it impacts a limited subset of features rather than the entire dimension. Consequently, feature modification is a subtler method.} In addition, modifying partial but not all the node features can preserve the performance of the main task as much as possible. The proposed NA$^{2}$ follows three stages: (i) Manipulate the local data according to neuron testing results (Section~\ref{dm}); (ii) Build the shadow model by using the manipulated data (Section~\ref{sm}); (iii) Generate the adversarial attacks via the shadow model (Section~\ref{aa}). The overview of NA$^2$ is shown in Fig.~\ref{fig:framwork} and the pseudocode is given in Algorithm~\ref{alo} in the appendix. We detail each step in the following subsections.

\subsection{Data Manipulation}
To obtain the candidate nodes and features to be modified, the data manipulation is divided into two steps: (i) Locate the significant neural path and obtain the candidate nodes. (ii) Select and modify the target features of the candidate nodes.
\label{dm}

\subsubsection{Locate Significant Neural Path}
NA$^{2}$ commits to locating significant neural paths related to the main task and manipulating the corresponding data, which curbs the performance degradation of the main task as much as possible. First, taking a $L$-layers fully-connected DNN as an example, we give the definition of the neural path as follows.

\emph{\textcolor{black}{\textbf{Definition 1(Significant Neuron Path):}} The neural path starts from the input neuron, \textcolor{black}{traverses the hidden layer,} and ends with the neuron in the output layer. The \textcolor{black}{significant} neural path $P$ can be expressed as:}
\textcolor{black}{\begin{equation}
    P = [k^{1}, k^{2}, ..., k^{L}],
\end{equation}}
\emph{\textcolor{black}{where $k^{1},k^{2},..,k^{L}$ are the  any neuron in the $1$-st, $2$-nd, ..., $L$-th layer of the DNN, respectively. The neurons in all other layers, except for $k^{L}$, are computed based on the neurons from the previous layer. The specific calculation process is demonstrated by Equation~\ref{k1st} and~\ref{kother}. }}

\emph{\textcolor{black}{\textbf{Definition 2(Output of the l-th hidden layer in GCN):}} Then, we introduce the neurons in GNN for a better understanding of NA$^{2}$. Take graph convolutional network (GCN)~\cite{kipf2016semi} as an example, the output of the $l$-th hidden layer (without activation function) can be described as:} 
\begin{equation}
\label{gcn}
    Z^l = H^{l+1} =\Tilde{D}^{-1/2}(A+I_{N})\Tilde{D}^{-1/2}H^{l}W^{l},
\end{equation}
\emph{where $\Tilde{D}$ is the degree matrix of $A+I_{N}$, $I_{N}$ is an identity matrix and $H^{0}=X$. Denote $W\in \mathbb{R}^{N\times M}$, where $N$ is the number of nodes in the graph and $M$ is the number of neurons in the $l$-th hidden layer. Thus, the output of the $i$-th neuron for a node $t$ is the $i$-th column of $Z^l_t$, which is denoted as $[Z^l_t]_{i}$.}

In order to locate the significant neurons for node classification, this process is performed after training for a number of epochs $\tau$, at which time the corresponding relationship between neurons and downstream tasks is established. More details about the number of epochs $\tau$ will be discussed later. Then, for a $L$-layers GNN, the first neuron of the node $t$ is started with the output layer:
\begin{equation}
\label{k1st}
    k^{l} =\underset{k^l}{\mathop{\arg \max }}[|Z_{t}^{l}|]_{k^l},
\end{equation}
where $|\cdot|$ is the absolute value operator. \textcolor{black}{The $argmax()$ itself is numerically stable because it only performs an index lookup operation, without involving rounding errors in numerical calculations. Even if the elements in the input vector are very close or equal, the function can always accurately return the index of the maximum value.}

\textcolor{black}{All neurons on this the significant neuron path are in an active state, and the activation of the end neuron is caused by the activation and ``contribution''  of other neurons on the path layer by layer. Therefore, we use partial derivatives to measure the ``contribution'' and lock the key neurons in other layers in reverse order.} Thus the rest significant neurons are selected by the following rule: 
\begin{equation}
\label{kother}
    k^{l-1} = \underset{k^{l-1}}{\mathop{\arg \max }} \Big[\Big |\frac{\partial[ Z_{t}^{l}]_{{k}^{l}}}{\partial Z_{t}^{l-1}} \cdot Z_{t}^{l-1}\Big | \Big]_{k^{l-1}}.
\end{equation}

 NA$^{2}$ traverses all layers of the local GNN according to the above rule, and finally obtains the node $t$'s reverse neural path $P_{t}=[k^{l},k^{l-1},...,k^{1}]$. Next, the train set is divided according to the node categories. For each category of nodes, we perform neuron tests on them and count the number of nodes with the same neural path. Then, take the neural path with the most nodes as the target path $P_T$, and add the rest nodes corresponding to other paths into the candidate nodes set $C$. \textcolor{black}{We use the ReLU function as the activation function. The derivative of ReLU is 1 in the positive interval and 0 in the negative interval. Therefore, the derivative of ReLU is numerically stable and does not exhibit abrupt numerical changes, which prevents the introduction of numerical errors during computation. The stability of the derivative of ReLU ensures that each step in the application of the chain rule is stable, so the process of calculating the partial derivatives in Equation~\ref{kother} is stable.}

\subsubsection{Select and Modify the Target Feature}

After we have obtained the significant neuron path $P_T$, we can identify the significant neuron of the local GNN's input layer $k^{1}$ by $P_T$. Then we extract the $k^{1}$-th column of the weight matrix $W^{1}$ in the first layer of the local GNN, which is denoted as $w_{k}^1$. The indexes of the top-$\textcolor{black}{B}$ of the $w_{k}^1$ are selected within the budget $\textcolor{black}{B}=\gamma\cdot d$ (i.e., the number of features to modify), and they are regarded as the target feature set $I$. $\gamma$ and $d$ are the scale of features modification and the number of node features, respectively.  At last, the $i$-th dimension of the candidate nodes' features are modified as:
\begin{equation}
\label{mod_fea}
\begin{split}
    &\Tilde{X}_{c}^{i} = \max (X_{c})\\
    s.t. \quad &c \in C \quad and \quad i \in I.
\end{split}
\end{equation}

The target features are replaced with the maximum value of the original features $X_{c}$, since we believe it maximizes the impact of the examples on the neural path.

\textcolor{black}{\textbf{Case Study:} Equation~\ref{mod_fea} uses the maximum value for replacement, which might ignore the possibility of outliers. To alleviate this concern, we conducted a comparative experiment. In this work, the SGA is adopted as the perturbation generator, we calculated the variance of all embeddings that need to be modified in the step of modifying node embeddings, and then did not modify the top 5\% of the embeddings (called NA2-FGA-N). TABLE~\ref{tab:my-table} records its attack results compared to normal attacks (called NA2-FGA), and the best attack performance highlighted in bold. It is clearly seen that the attack of NA2-FGA is better than NA2-FGA-N on all datasets, indicating that our method remains effective under high variance.}

\begin{table}[ht]
    \centering
    \caption{\textcolor{black}{Comparison of attack success rate between NA2-FGA and NA2-FGA-N for GCN local model.}}
    \label{tab:my-table}
    \begin{tabular}{c|c|cc}
        \hline
        \multirow{2}{*}{Local model} & \multirow{2}{*}{Dataset} & \multicolumn{2}{c}{Attack Method} \\
        \cline{3-4} 
         &  & NA2-FGA & NA2-FGA-N \\
        \hline
        \multirow{5}{*}{GCN} & Cora & \textbf{52\% }& 49\% \\
        \cline{2-4} 
         & Cora ML & \textbf{67\%} & 63\% \\
        \cline{2-4} 
         & Citeseer & \textbf{86\%} & 80\% \\
        \cline{2-4} 
         & Pubmed & \textbf{90\%} & 84\% \\
        \cline{2-4} 
         & ogbn-arxiv & \textbf{68\%} & 66\% \\
        \hline
    \end{tabular}
\end{table}

 \textcolor{black}{It can be inferred from the above formulas and analysis that neuron testing is a core component of the NA$^{2}$. It analyzes neuron activation to help us identify the key features that have the greatest impact on model decisions. First, neuron testing identifies neural paths closely related to the target task and further screens out the feature subset that has the greatest impact on the key neurons. These features become the targets of data manipulation. By modifying these key features, we can more effectively influence model decisions and thus improve the attack success rate. In addition, neuron testing helps to narrow the scope of data manipulation, avoiding modification of a large number of irrelevant features, thereby improving attack efficiency and reducing the impact on the performance of normal tasks, as well as the risk of being detected. Finally, neuron testing helps us understand the similarity between different model structures, thereby supporting transfer attacks. Even without knowing the structure of the target model, effective attacks can still be carried out.}

\subsection{Shadow Model Construction}
\label{sm}

\textcolor{black}{The shadow model refers to a model with similar behavior to the server model, constructed by the malicious client using its own manipulated data and the prediction probabilities returned by the server model. The shadow model acts as a substitute for the server model. It allows the malicious client to generate adversarial embeddings without repeatedly querying the server model, which is both efficient and reduces the risk of detection in the VFGL scenario.} First, we construct the shadow model $\Tilde{\mathcal{S}}$ by cascading the local GNN and a $L$-layers MLP. Then we query for probabilities $p$ which are used to train the shadow model. The probabilities encapsulate both the main task's essential information and the server model's behavior. Thus, we consider the mean square error (MSE) as the target of the shadow model during the training process:
\begin{equation}
\label{loss_shadow}
    \mathcal{L}_{\Tilde{\mathcal{S}}}=\frac{1}{R\times |F|}\sum^{R}_{i=1}\sum^{|F|}_{j=1}(\Tilde{\mathcal{S}}(A,\Tilde{X})-p_{i,j})^{2},
\end{equation}
where $R$ is the number of nodes for training the shadow model, and $|F|$ is the number of class for nodes. \textcolor{black}{Since MSE is a convex function and continuously differentiable, its minimum value is the global optimal solution. We can use methods such as gradient descent to minimize MSE, thereby finding the best parameters for the shadow model. Therefore, the optimization process is convergent.}

Then we explain why the MSE can be applied to guide the shadow model to mimic the behavior of the server model in VFGL. First, we give the definition of the decision boundary.

\emph{\textbf{Definition 3:} The decision boundary of a classifier $\mathcal{V}$ can be expressed as follows:}
\begin{equation}
\begin{split}
    B_{\mathcal{V}}:\quad &\mathcal{V}_{l}(x)-\mathcal{V}_{\hat{l}}(x)=0\\
    s.t.& \quad \hat{l}=\underset{l}\arg \max \mathcal{V}(x),
\end{split}
\end{equation}
\emph{where $l$ is the $l$-th class of probabilities, and $x$ is the input of the classifier.}

Then, the objective of the MSE can be described as:
\begin{equation}
\begin{split}
    &\underset{\Tilde{\mathcal{S}}}{\arg \min} MSE(\Tilde{\mathcal{S}}(A, \Tilde{X}), \mathcal{S}(h_{global}))\\
    &\Leftrightarrow \underset{\Tilde{\mathcal{S}}}{\arg \min} |\Tilde{\mathcal{S}}(A, \Tilde{X})- \mathcal{S}(h_{global})|\\
    &\Leftrightarrow \underset{\Tilde{\mathcal{S}}}{\arg \min} |(\Tilde{\mathcal{S}}_{l}(A, \Tilde{X})-\Tilde{\mathcal{S}}_{\hat{l}}(A, \Tilde{X}))\\
    &\quad \quad -(\mathcal{S}_{l}(h_{global})-\mathcal{S}_{\hat{l}}(h_{global}))|\\
    &\Leftrightarrow \underset{\Tilde{\mathcal{S}}}{\arg \min} |B_{\Tilde{\mathcal{S}}}-B_{\mathcal{S}}|.
\end{split}
\end{equation}

Therefore, using MSE is helpful in fitting the shadow model to the server model. A shadow model that is more similar to the server model can provide more accurate information to the attackers.

\subsection{Adversarial Attacks}
\label{aa}
The shadow model constructed in the above steps can be used for facilitating various centralized attackers to improve the attack success rate. In this work, we consider generating adversarial perturbations by adding/deleting edges in the graph, which is mainstream and effective. The adversarial embeddings for the target node $v_t$ produced by the malicious client $m$ are formulated as:
\begin{equation}
\label{ad_emb}
\begin{split}
    &\hat{h}^{v_{t}}_{m}={f}^{m}_{\theta}(\hat{A}, \Tilde{X}, v_{t})\\
    &s.t. \quad \hat{A}=ATK(A,\Tilde{X},\mathcal{\Tilde{S}}),
\end{split}
\end{equation}
where $ATK(\cdot)$ is a kind of centralized attack method. And the perturbations are limited within the attack budget $\Delta$, i.e., $||\hat{A}-A||_0\leq 2\cdot \Delta$. It should be noted that $A$ and $\hat{A}$ are symmetrical in the undirected graph.

\section{Experiments}
\label{experiments}
In this section, we conduct experiments to evaluate the effectiveness of NA$^2$ and four research questions (RQs) are to be answered:
\begin{itemize}
    \item \textbf{RQ1:} Can NA$^{2}$ improve the attacks' performance significantly? What is the query cost of NA$^{2}$?
    \item \textbf{RQ2:} Is there a close relationship between the attack success rate and the contribution of the adversary?
    \item \textbf{RQ3:} Can NA$^{2}$ still work well in defensive VFGL?
    \item \textbf{RQ4:} How do the scale of modified features and the round of modification affect NA$^2$? What is the time complexity?
\end{itemize}
\subsection{Experimental Settings}
\subsubsection{Datasets}
The proposed method is evaluated on \textcolor{black}{five} real-world datasets: Cora~\cite{mccallum2000automating}, Cora\_ML~\cite{mccallum2000automating}, Citeseer~\cite{mccallum2000automating}, Pubmed~\cite{sen2008collective} and \textcolor{black}{ogbn-arxiv~\cite{hu2020open}}, the information of which is shown in TABLE~\ref{Datasets}. 


To simulate the VFGL learning scenario where clients may not have the same data features, we divide the data and assign them to the clients. 
However, the graphs in these datasets are sparse, i.e., the average degree is low as shown in TABLE~\ref{Datasets}. To avoid the appearance of isolated nodes after segmenting the graph, we randomly segment the features only while keeping the edges the same for every client. We test the performance of NA$^{2}$ 5 times as well as other baselines and report the average results to eliminate the impact of the randomness. 

\begin{table}[htpb]
    \centering
    \caption{\textcolor{black}{The basic statistics of five graph-structured datasets.}}
    \label{Datasets}
    \resizebox{0.9\linewidth}{!}{\LARGE
    \begin{tabular}{c|ccccc}
    \hline \hline
    \textbf{Dataset}&\textbf{\#Nodes}&\textbf{\#Edges}&\textbf{\#Features}&\textbf{\#Classes}&\textbf{\#Average Degree}\\
    \hline
         \textbf{Cora~\cite{mccallum2000automating}}&2,708&5,429&1,433&7&2.00\\
         \textbf{Cora\_ML~\cite{mccallum2000automating}}&2,810&7,981&2,879&7&2.84\\
         \textbf{Citeseer~\cite{mccallum2000automating}}&3,327&4,732&3,703&6&1.42\\
         \textbf{Pubmed~\cite{sen2008collective}}&19,717&44,325&500&3&2.25\\
         \textbf{ogbn-arxiv~\cite{hu2020open}}&16,9343&1,166,243&128&40&13.7\\
    \hline \hline
    \end{tabular}}
\end{table}

\subsubsection{Local \textcolor{black}{and Shadow Model}}
Three kinds of GNNs are adopted as the local \textcolor{black}{and shadow model} in VFGL.
\begin{itemize}
    \item \textbf{GCN~\cite{kipf2016semi}:} For node classification, a two-layer GCN is applied as the local model in the VFGL. \textcolor{black}{According to Equation~\ref{gcn}, the hidden node representation of $(l+1)$-th layer is expressed as:}
    \begin{equation}
        \textcolor{black}{H^{l+1} = \sigma(\Tilde{D}^{-1/2}\Tilde{A}\Tilde{D}^{-1/2}H^{l}W^{l}),}
    \end{equation}
    \textcolor{black}{where $\sigma$ is the activation function, e.g., ReLU. And $\Tilde{A}=A+I_{N}$, $I_{N}$ is unit matrix.} 
    \item \textbf{SGC~\cite{wu2019simplifying}:} SGC removes the nonlinear activation function of GCN, \textcolor{black}{and the forward of SGC is expressed as:}
    \begin{equation}
        \textcolor{black}{H^{l+1}=\Tilde{S^{l}}X\Tilde{W},}
    \end{equation}
    \textcolor{black}{where $\Tilde{S^{l}}=\Tilde{A}\Tilde{A}...\Tilde{A}$ depends on the layers of SGC, and $\Tilde{W}=W^{0}W^{1}...W^{l}$ denotes a collapsed weight matrix.}
    \item \textbf{GCNII~\cite{chen2020simple}:} To address the over-smoothing problem, GCNII adopts initial residual connection and identity mapping, \textcolor{black}{and the node representation is described as:}
    \begin{equation}
    \scriptsize
        \textcolor{black}{H^{l+1}=\sigma(((1-\alpha_{l})\Tilde{A}H^{l}+\alpha_{l}H^{0})((1-\beta_{l})I_{N}+\beta_{l}W^{l})),}
    \end{equation}
    \textcolor{black}{where $\alpha_{l}$ and $\beta_{l}$ are hyperparameters to control the scale of input features and adaptive weight decay, respectively. $I_{N}$ is identity matrix for identity mapping.}
\end{itemize}

\subsubsection{Evaluation Metrics}
To measure the proposed method, attack success rate (ASR), contribution score (CS) and average queries (AQ) are adopted to evaluate the effectiveness and efficiency of NA$^{2}$.
\begin{itemize}
    \item \textbf{ASR:} The ASR is defined as follows:
    \begin{equation}
        ASR = \frac{N_{s}}{N_{a}}\times 100\%,
    \end{equation}
    where $N_{s}$ is the number of nodes which are attacked successfully, and $N_{a}$ is the number of target nodes.
    \item \textbf{CS:} Similar to~\cite{lyu2020collaborative}, it is a metric to measure the contribution of each client during the training process. The CS of client $i$ is defined as:
    \begin{equation}
    \begin{split}
        &CS_{i}=\frac{con_{i}}{\sum^{K}_{k}con_{k}}\\
        s.t. \quad &con_{i} = sinh(\alpha\cdot \frac{acc_{i}}{\sum^{K}_{k}acc_{k}}),
    \end{split}
    \end{equation}
    where $acc_{i}$ is the training accuracy of the server model when only $i$'s data is input. $\alpha$ is the scaling constant of contribution. In this work, it is set to 5. $K$ is the number of clients in the VFGL.
    \item \textbf{AQ:} The AQ is applied to measure the efficiency of the attack, which is defined as:
    \begin{equation}
        AQ=\frac{1}{N_{a}}\sum^{N_{a}}_{j}q_{j}.
    \end{equation}
    
    If the attack fails, the number of queries $q$ is set to a default query \textcolor{black}{budget} $Q$. It should be pointed out that we take the number of times that a malicious client initiates a request to the server model as the number of queries. Frequent requests to the server in a short period of time are suspicious and inefficient. Therefore, the lower the AQ, the higher the efficiency of the attacker.
\end{itemize}

\begin{table}
    \centering
    \caption{The structure and parameters of the local GNN.}
    \label{settings}
    \resizebox{0.9\linewidth}{!}{\LARGE
    \begin{tabular}{c|cccc}
    \hline \hline
    \textbf{Model}&\textbf{\#Layers}&\textbf{\#Hidden dims}&\textbf{\#Out dims}&\textbf{\#Training period}\\
    \hline
         \textbf{GCN}&2&32&16&200\\
         \textbf{SGC}&2&32&16&200\\
         \textbf{GCNII}&4&32, 32, 32&16&1,000\\
    \hline \hline
    \end{tabular}}
\end{table}

\subsubsection{Baselines and Adversarial Attacks}
\emph{\textbf{Baselines.}}
Due to the lack of research, the proposed NA$^{2}$ is compared with the other two methods we propose: 
\begin{itemize}
    \item \textbf{Random Features Attack (RFA):} It randomly selects the same number of features as NA$^{2}$ for each candidate node, and the value is set to the largest value in the original feature distribution.
    \item \textbf{Specific Features Attack (SFA):} It randomly selects the same number of features as the specific features set. For each candidate node, the specific features are set to the largest value in the original feature distribution.
\end{itemize}

For a fair comparison, the number of candidate nodes is the same as NA$^{2}$.

\emph{\textbf{Adversarial attacks.}} We consider four advanced centralized adversarial attacks \textbf{FGA~\cite{chen2018fast},} \textbf{GradArgmax~\cite{dai2018adversarial}, }
\textbf{Nettack~\cite{zugner2018adversarial},}
\textbf{SGA~\cite{li2021adversarial}}
as the attack generator. Moreover, to illustrate the query efficiency of NA$^{2}$, we select a query-based black-box adversarial attack method for comparison. They are briefly described in the appendix B.


\subsubsection{Parameter Settings}
For different GNNs, the structure and parameters are summarized in TABLE~\ref{settings}. The VFGL is trained using Adam with the learning rate set to 0.01. ReLU is adopted as the activation function for GCN and GCNII. For the malicious client, the scale $\gamma$ for feature modification is searched in [0.01, 0.1], i.e., only $\gamma \cdot d$ ($d$ is the number of node features) features are allowed to be modified per node. For the perturbation generator, the attack budget $\Delta$ is set to 1, considering the sparsity of the graph.

\subsubsection{Experimental Environment}
Our experimental environment consists of Intel XEON 6240 2.6GHz x 18C (CPU), Tesla V100 32GiB (GPU), 16GiB memory (DDR4-RECC 2666) and Ubuntu 16.04 (OS).

\begin{table*}
\renewcommand\arraystretch{0.6}
\centering
\caption{\textcolor{black}{The average attack success rate ($\pm$ std) and improvement of four adversarial attacks on three kinds of local GNNs.}}
\resizebox{\textwidth}{!}{
\begin{tabular}{c|c|c|cccccccc}
\hline \hline
                                       &                                     &                                   & \multicolumn{8}{c}{\textbf{Attack Method}}                                                                                                                                                                                                                                                                                                                        \\ \cline{4-11} 
                                       &                                     &                                   & \multicolumn{2}{c}{\textbf{FGA}}                                                       & \multicolumn{2}{c}{\textbf{GradArgmax}}                                                & \multicolumn{2}{c}{\textbf{Nettack}}                                                   & \multicolumn{2}{c}{\textbf{SGA}}                                                       \\ \cline{4-11} 
\multirow{-3}{*}{\textbf{Local Model}} & \multirow{-3}{*}{\textbf{Dataset}}  & \multirow{-3}{*}{\textbf{Method}} & \textbf{ASR (\%)}                            & \textbf{Impv (\%)}                   & \textbf{ASR (\%)}                            & \textbf{Impv (\%)}                   & \textbf{ASR (\%)}                            & \textbf{Impv (\%)}                   & \textbf{ASR (\%)}                            & \textbf{Impv (\%)}                   \\ \hline
                                       &                                     & \textbf{Clean}                    & 17.15±2.69                                   & /                                       & 8.18±2.57                                    & /                                       & 22.55±3.06                                   & /                                       & 16.59±2.66                                   & /                                       \\
                                       &                                     & \textbf{RFA}                  & 23.20±4.55                                   & 35.28                                   & 7.15±3.80                                    & -12.59                                  & 28.11±4.09                                   & 24.66                                   & 22.52±3.19                                   & 35.74                                   \\
                                       &                                     & \textbf{SFA}                  & 18.01±1.35                                   & 5.01                                    & 7.83±3.59                                    & -4.28                                   & 22.06±2.76                                   & -2.17                                   & 16.64±2.70                                   & 0.30                                    \\
                                       & \multirow{-4}{*}{\textbf{Cora}}     & \textbf{NA$^{2}$}                      & \cellcolor[HTML]{FFCCCC}\textbf{52.37±6.43}  & \cellcolor[HTML]{FFCCCC}\textbf{205.36} & \cellcolor[HTML]{FFCCCC}\textbf{35.55±13.72} & \cellcolor[HTML]{FFCCCC}\textbf{334.60} & \cellcolor[HTML]{FFCCCC}\textbf{63.35±7.05}  & \cellcolor[HTML]{FFCCCC}\textbf{180.93} & \cellcolor[HTML]{FFCCCC}\textbf{45.01±8.10}  & \cellcolor[HTML]{FFCCCC}\textbf{171.31} \\ \cline{2-11} 
                                       &                                     & \textbf{Clean}                    & 12.58±4.83                                   & /                                       & 10.95±5.86                                   & /                                       & 16.24±6.09                                   & /                                       & 10.43±2.88                                   & /                                       \\
                                       &                                     & \textbf{RFA}                  & 23.88±2.74                                   & 89.89                                   & 16.24±6.09                                   & 48.36                                   & 18.60±6.67                                   & 17.53                                   & 13.59±3.18                                   & 30.32                                   \\
                                       &                                     & \textbf{SFA}                  & 22.06±4.07                                   & 75.41                                   & 12.93±6.35                                   & 18.13                                   & 21.97±5.37                                   & 18.12                                   & 13.78±2.18                                   & 32.14                                   \\
                                       & \multirow{-4}{*}{\textbf{Cora\_ML}} & \textbf{NA$^{2}$}                      & \cellcolor[HTML]{FFCCCC}\textbf{67.34±6.30}  & \cellcolor[HTML]{FFCCCC}\textbf{435.50} & \cellcolor[HTML]{FFCCCC}\textbf{40.90±17.00} & \cellcolor[HTML]{FFCCCC}\textbf{273.67} & \cellcolor[HTML]{FFCCCC}\textbf{63.90±13.67} & \cellcolor[HTML]{FFCCCC}\textbf{243.55} & \cellcolor[HTML]{FFCCCC}\textbf{38.98±14.31} & \cellcolor[HTML]{FFCCCC}\textbf{273.80} \\ \cline{2-11} 
                                       &                                     & \textbf{Clean}                    & 28.53±7.08                                   & /                                       & 17.12±5.30                                   & /                                       & 30.56±3.84                                   & /                                       & 23.94±2.91                                   & /                                       \\
                                       &                                     & \textbf{RFA}                  & 58.97±12.17                                  & 106.69                                  & 23.68±10.03                                  & 38.32                                   & 63.34±12.76                                  & 107.26                                  & 45.17±10.03                                  & 88.68                                   \\
                                       &                                     & \textbf{SFA}                  & 28.58±6.01                                   & 0.18                                    & 14.82±6.12                                   & -13.43                                  & 29.82±4.30                                   & -2.42                                   & 19.63±4.91                                   & -18.00                                  \\
                                       & \multirow{-4}{*}{\textbf{Citeseer}} & \textbf{NA$^{2}$}                      & \cellcolor[HTML]{FFCCCC}\textbf{86.30±4.48}  & \cellcolor[HTML]{FFCCCC}\textbf{202.49} & \cellcolor[HTML]{FFCCCC}\textbf{41.13±13.27} & \cellcolor[HTML]{FFCCCC}\textbf{140.25} & \cellcolor[HTML]{FFCCCC}\textbf{87.30±2.45}  & \cellcolor[HTML]{FFCCCC}\textbf{185.67} & \cellcolor[HTML]{FFCCCC}\textbf{73.36±8.80}  & \cellcolor[HTML]{FFCCCC}\textbf{206.43} \\ \cline{2-11} 
                                       &                                     & \textbf{Clean}                    & 52.50±18.77                                  & /                                       & 38.97±24.69                                  & /                                       & 48.94±20.75                                  & /                                       & 52.74±18.32                                  & /                                       \\
                                       &                                     & \textbf{RFA}                  & 63.45±6.85                                   & 20.86                                   & 37.56±21.16                                  & -3.61                                   & 55.86±16.25                                  & 14.16                                   & 58.45±12.49                                            &  10.83                                       \\
                                       &                                     & \textbf{SFA}                  & 58.38±18.43                                  & 11.21                                   & 30.82±14.53                                  & -20.92                                  & 49.23±18.35                                  & 0.60                                    & 54.07±17.20                                  & 2.52                                    \\
                                       & \multirow{-4}{*}{\textbf{Pubmed}}   & \textbf{NA$^{2}$}                      & \cellcolor[HTML]{FFCCCC}\textbf{89.90±9.22}  & \cellcolor[HTML]{FFCCCC}\textbf{65.53}  & \cellcolor[HTML]{FFCCCC}\textbf{60.71±16.99} & \cellcolor[HTML]{FFCCCC}\textbf{55.79}  & \cellcolor[HTML]{FFCCCC}\textbf{83.47±17.69} & \cellcolor[HTML]{FFCCCC}\textbf{70.56}  & \cellcolor[HTML]{FFCCCC}\textbf{77.28±17.45} & \cellcolor[HTML]{FFCCCC}\textbf{51.42}  \\ \cline{2-11}
                                       &                                     & \textbf{Clean}                    & 33.66±2.25                                   & /                                       & 25.65±4.87                                   & /                                       & 35.68±2.96                                   & /                                       & 39.58±4.32                                   & /                                       \\
                                       &                                     & \textbf{RFA}                  & 28.36±4.23                                  & -15.75                                 & 28.36±4.58                                  & 10.57                                  & 44.24±3.62                                 & 23.99                                  & 41.23±4.51                                  & 4.17                                  \\
                                       &                                     & \textbf{SFA}                  & 35.87±2.58                                  & -6.57                                   & 29.58±2.45                                  & 15.32                                 & 35.35±6.48                                  & -0.92                                   & 51.63±4.83                                   & 30.44                                 \\
\multirow{-20}{*}{\textbf{GCN}}        & \multirow{-4}{*}{\textbf{ogbn-arxiv}} & \textbf{NA$^{2}$}                      & \cellcolor[HTML]{FFCCCC}\textbf{68.24±4.58}  & \cellcolor[HTML]{FFCCCC}\textbf{102.73} & \cellcolor[HTML]{FFCCCC}\textbf{40.86±5.46} & \cellcolor[HTML]{FFCCCC}\textbf{59.30} & \cellcolor[HTML]{FFCCCC}\textbf{69.56±6.41}  & \cellcolor[HTML]{FFCCCC}\textbf{94.96} & \cellcolor[HTML]{FFCCCC}\textbf{64.25±2.36}  & \cellcolor[HTML]{FFCCCC}\textbf{62.33} \\ \hline 
                                       &                                     & \textbf{Clean}                    & 16.27±1.94                                   & /                                       & 7.42±5.12                                    & /                                       & 23.49±4.95                                   & /                                       & 17.84±1.93                                   & /                                       \\
                                       &                                     & \textbf{RFA}                  & 24.01±5.48                                   & 47.60                                   & 10.96±5.37                                   & 47.84                                   & 25.85±2.91                                   & 10.02                                   & 23.67±2.12                                   & 32.69                                   \\
                                       &                                     & \textbf{SFA}                  & 17.22±3.62                                   & 5.89                                    & 7.53±4.34                                    & 1.59                                    & 22.46±4.29                                   & -4.42                                   & 17.96±2.25                                   & 0.68                                    \\
                                       & \multirow{-4}{*}{\textbf{Cora}}     & \textbf{NA$^{2}$}                      & \cellcolor[HTML]{FFCCCC}\textbf{46.41±14.50} & \cellcolor[HTML]{FFCCCC}\textbf{185.31} & \cellcolor[HTML]{FFCCCC}\textbf{22.42±14.67} & \cellcolor[HTML]{FFCCCC}\textbf{202.32} & \cellcolor[HTML]{FFCCCC}\textbf{57.26±12.55} & \cellcolor[HTML]{FFCCCC}\textbf{143.70} & \cellcolor[HTML]{FFCCCC}\textbf{41.43±11.46} & \cellcolor[HTML]{FFCCCC}\textbf{132.26} \\ \cline{2-11} 
                                       &                                     & \textbf{Clean}                    & 22.06±4.71                                   & /                                       & 19.21±5.06                                   & /                                       & 22.46±3.16                                   & /                                       & 12.78±3.10                                   & /                                       \\
                                       &                                     & \textbf{RFA}                  & 41.69±9.82                                   & 89.00                                   & 27.38±12.04                                  & 42.54                                   & 40.75±8.83                                   & 81.41                                   & 23.35±4.18                                   & 82.70                                   \\
                                       &                                     & \textbf{SFA}                  & 20.73±4.40                                   & -17.08                                  & 17.51±4.57                                   & -11.85                                  & 21.44±2.54                                   & 6.56                                    & 13.19±3.18                                   & -5.69                                   \\
                                       & \multirow{-4}{*}{\textbf{Cora\_ML}} & \textbf{NA$^{2}$}                      & \cellcolor[HTML]{FFCCCC}\textbf{80.93±6.31}  & \cellcolor[HTML]{FFCCCC}\textbf{266.87} & \cellcolor[HTML]{FFCCCC}\textbf{67.33±12.02} & \cellcolor[HTML]{FFCCCC}\textbf{250.56} & \cellcolor[HTML]{FFCCCC}\textbf{86.54±3.81}  & \cellcolor[HTML]{FFCCCC}\textbf{276.32} & \cellcolor[HTML]{FFCCCC}\textbf{50.11±15.21} & \cellcolor[HTML]{FFCCCC}\textbf{292.16} \\ \cline{2-11} 
                                       &                                     & \textbf{Clean}                    & 33.17±5.20                                   & /                                       & 22.33±5.63                                   & /                                       & 33.15±5.87                                   & /                                       & 23.60±2.19                                   & /                                       \\
                                       &                                     & \textbf{RFA}                  & 62.91±12.26                                  & 110.79                                  & 29.56±11.01                                  & 32.37                                   & 76.75±9.71                                   & 131.53                                  & 58.38±11.12                                  & 147.31                                  \\
                                       &                                     & \textbf{SFA}                  & 27.50±8.23                                   & -6.05                                   & 19.69±8.72                                   & -8.85                                   & 35.32±7.03                                   & -4.56                                   & 22.26±4.62                                   & 3.24                                    \\
                                       & \multirow{-4}{*}{\textbf{Citeseer}} & \textbf{NA$^{2}$}                      & \cellcolor[HTML]{FFCCCC}\textbf{76.36±15.60} & \cellcolor[HTML]{FFCCCC}\textbf{130.22} & \cellcolor[HTML]{FFCCCC}\textbf{44.22±23.85} & \cellcolor[HTML]{FFCCCC}\textbf{98.01}  & \cellcolor[HTML]{FFCCCC}\textbf{85.00±7.42}  & \cellcolor[HTML]{FFCCCC}\textbf{156.41} & \cellcolor[HTML]{FFCCCC}\textbf{61.29±9.63}  & \cellcolor[HTML]{FFCCCC}\textbf{159.65} \\ \cline{2-11} 
                                       &                                     & \textbf{Clean}                    & 56.68±19.94                                  & /                                       & 24.76±22.65                                  & /                                       & 53.95±21.09                                  & /                                       & 52.38±12.58                                  & /                                       \\
                                       &                                     & \textbf{RFA}                  & 60.40±13.36                                  & 6.56                                    & 26.52±17.49                                  & 7.11                                    & 53.59±22.64                                  & -0.68                                   & 59.87±11.60                                 & 14.30                                   \\
                                       &                                     & \textbf{SFA}                  & 64.20±7.00                                   & 13.27                                   & 24.10±13.97                                  & -2.65                                   & 50.61±21.29                                  & -6.21                                   & 52.00±11.59                                  & -0.73                                   \\
                                       & \multirow{-4}{*}{\textbf{Pubmed}}   & \textbf{NA$^{2}$}                      & \cellcolor[HTML]{FFCCCC}\textbf{76.55±24.88} & \cellcolor[HTML]{FFCCCC}\textbf{35.04}  & \cellcolor[HTML]{FFCCCC}\textbf{54.11±20.00} & \cellcolor[HTML]{FFCCCC}\textbf{118.52} & \cellcolor[HTML]{FFCCCC}\textbf{79.79±28.70} & \cellcolor[HTML]{FFCCCC}\textbf{47.89}  & \cellcolor[HTML]{FFCCCC}\textbf{73.62±27.30} & \cellcolor[HTML]{FFCCCC}\textbf{40.54}  \\ \cline{2-11}
                                       &                                     & \textbf{Clean}                    & 46.52±13.54                                   & /                                       & 28.79±20.31                                   & /                                       & 49.67±19.58                                   & /                                       & 45.20±22.69                                   & /                                       \\
                                       &                                     & \textbf{RFA}                  & 49.57±8.45                                  & 6.56                                  & 27.45±6.35                                  & -4.65                                   & 55.68±14.56                                  & 12.10                                 & 49.64±5.96                                  & 9.82                                   \\
                                       &                                     & \textbf{SFA}                  & 51.36±7.44                                  & 10.40                                    & 31.25±5.47                                   & 8.54                                  & 54.63±8.47                                   & 9.99                                   & 38.54±5.22                                   & -14.73                                  \\
\multirow{-20}{*}{\textbf{SGC}}        & \multirow{-4}{*}{\textbf{ogbn-arxiv}} & \textbf{NA$^{2}$}                      & \cellcolor[HTML]{FFCCCC}\textbf{68.21±24.80}  & \cellcolor[HTML]{FFCCCC}\textbf{46.63} & \cellcolor[HTML]{FFCCCC}\textbf{42.45±15.41} & \cellcolor[HTML]{FFCCCC}\textbf{47.45} & \cellcolor[HTML]{FFCCCC}\textbf{63.48±23.85}  & \cellcolor[HTML]{FFCCCC}\textbf{27.8} & \cellcolor[HTML]{FFCCCC}\textbf{58.65±19.42}  & \cellcolor[HTML]{FFCCCC}\textbf{29.76} \\ \hline
                                       &                                     & \textbf{Clean}                    & 33.59±14.57                                  & /                                       & 28.06±19.28                                  & /                                       & 46.89±23.58                                  & /                                       & 24.72±5.87                                   & /                                       \\
                                       &                                     & \textbf{RFA}                  & 37.24±16.54                                  & 10.88                                   & 21.83±16.40                                  & -22.20                                  & 52.60±22.35                                  & 12.16                                   & 37.26±2.24                                   & 50.73                                   \\
                                       &                                     & \textbf{SFA}                  & 28.66±14.87                                  & -14.66                                  & 20.46±15.71                                  & -27.08                                  & 36.96±24.53                                  & -21.19                                  & 20.08±4.51                                   & -18.77                                  \\
                                       & \multirow{-4}{*}{\textbf{Cora}}     & \textbf{NA$^{2}$}                      & \cellcolor[HTML]{FFCCCC}\textbf{62.01±6.33}  & \cellcolor[HTML]{FFCCCC}\textbf{84.64}  & \cellcolor[HTML]{FFCCCC}\textbf{31.78±14.59} & \cellcolor[HTML]{FFCCCC}\textbf{13.26}  & \cellcolor[HTML]{FFCCCC}\textbf{75.22±8.93}  & \cellcolor[HTML]{FFCCCC}\textbf{60.40}  & \cellcolor[HTML]{FFCCCC}\textbf{51.05±14.25} & \cellcolor[HTML]{FFCCCC}\textbf{106.51} \\ \cline{2-11} 
                                       &                                     & \textbf{Clean}                    & 30.39±9.96                                   & /                                       & 23.96±11.30                                  & /                                       & 33.64±9.69                                   & /                                       & 19.53±5.43                                   & /                                       \\
                                       &                                     & \textbf{RFA}                  & 55.24±17.80                                  & 81.79                                   & 34.03±20.11                                  & 54.22                                   & 61.41±13.60                                            & 82.55                                       & 38.39±10.43                                  & 96.59                                   \\
                                       &                                     & \textbf{SFA}                  & 28.60±9.54                                   & -5.87                                   & 23.73±9.85                                   & -0.94                                   & 36.06±12.51                                  & 7.21                                    & 17.65±4.55                                   & -9.64                                   \\
                                       & \multirow{-4}{*}{\textbf{Cora\_ML}} & \textbf{NA$^{2}$}                      & \cellcolor[HTML]{FFCCCC}\textbf{73.51±15.14} & \cellcolor[HTML]{FFCCCC}\textbf{141.90} & \cellcolor[HTML]{FFCCCC}\textbf{58.92±21.15} & \cellcolor[HTML]{FFCCCC}\textbf{145.93} & \cellcolor[HTML]{FFCCCC}\textbf{79.28±12.70} & \cellcolor[HTML]{FFCCCC}\textbf{135.67}  & \cellcolor[HTML]{FFCCCC}\textbf{39.58±14.01} & \cellcolor[HTML]{FFCCCC}\textbf{102.69} \\ \cline{2-11} 
                                       &                                     & \textbf{Clean}                    & 40.25±14.41                                  & /                                       & 17.21±14.04                                  & /                                       & 48.61±13.52                                  & /                                       & 37.77±4.36                                   & /                                       \\
                                       &                                     & \textbf{RFA}                  & 58.82±12.63                                  & 46.15                                   & 29.22±10.59                                  & 69.79                                   & 69.44±14.60                                  & 42.84                                   & 57.85±8.63                                   & 53.17                                   \\
                                       &                                     & \textbf{SFA}                  & 43.90±19.58                                  & 9.08                                    & 23.52±17.98                                  & 36.70                                   & 45.38±21.07                                  & -6.64                                   & 37.07±5.11                                   & -1.86                                   \\
                                       & \multirow{-4}{*}{\textbf{Citeseer}} & \textbf{NA$^{2}$}                      & \cellcolor[HTML]{FFCCCC}\textbf{69.11±11.53} & \cellcolor[HTML]{FFCCCC}\textbf{71.72}  & \cellcolor[HTML]{FFCCCC}\textbf{32.39±19.53} & \cellcolor[HTML]{FFCCCC}\textbf{88.23}  & \cellcolor[HTML]{FFCCCC}\textbf{78.34±20.46} & \cellcolor[HTML]{FFCCCC}\textbf{61.16}  & \cellcolor[HTML]{FFCCCC}\textbf{61.15±9.60}  & \cellcolor[HTML]{FFCCCC}\textbf{61.90}  \\ \cline{2-11} 
                                       &                                     & \textbf{Clean}                    & 56.39±18.05                                  & /                                       & 41.36±9.10                                   & /                                       & 49.26±13.82                                  & /                                       & 42.29±6.02                                   & /                                       \\
                                       &                                     & \textbf{RFA}                  & 47.88±11.01                                  & -15.09                                  & 32.45±13.45                                            &-21.60                                         & 44.76±21.20                                  & -9.14                                   & 42.79±6.91                                   & 1.18                                    \\
                                       &                                     & \textbf{SFA}                  & 55.92±12.30                                  & -0.84                                   & 40.10±7.98                                   & -3.14                                   & 56.16±10.98                                  & 14.01                                   & 45.96±4.22                                   & 8.66                                    \\
                                       & \multirow{-4}{*}{\textbf{Pubmed}}   & \textbf{NA$^{2}$}                      & \cellcolor[HTML]{FFCCCC}\textbf{87.84±8.47}  & \cellcolor[HTML]{FFCCCC}\textbf{55.77}  & \cellcolor[HTML]{FFCCCC}\textbf{59.44±14.44} & \cellcolor[HTML]{FFCCCC}\textbf{43.59}  & \cellcolor[HTML]{FFCCCC}\textbf{87.79±7.62}  & \cellcolor[HTML]{FFCCCC}\textbf{78.21}  & \cellcolor[HTML]{FFCCCC}\textbf{48.51±8.33}  & \cellcolor[HTML]{FFCCCC}\textbf{14.71}  \\ \cline{2-11} 
                                       &                                     & \textbf{Clean}                    & 50.84±19.48                                   & /                                       & 29.54±8.99                                   & /                                       & 35.58±16.25                                   & /                                       & 39.59±14.86                                   & /                                       \\
                                       &                                     & \textbf{RFA}                  & 54.36±9.58                                  & 6.92                                & 26.58±14.43                                 & -10.02                                   & 45.69±8.47                                 & 28.41                                  & 40.36±8.95                                  & 1.95                                 \\
                                       &                                     & \textbf{SFA}                  & 46.58±9.45                                   & -8.38                                    & 28.65±12.24                                   & -3.01                                 & 39.45±10.68                                  & 10.88                                   & 45.57±14.33                                   & 15.10                                 \\
\multirow{-20}{*}{\textbf{GCNII}}        & \multirow{-4}{*}{\textbf{ogbn-arxiv}} & \textbf{NA$^{2}$}                      & \cellcolor[HTML]{FFCCCC}\textbf{75.85±15.36}  & \cellcolor[HTML]{FFCCCC}\textbf{49.19} & \cellcolor[HTML]{FFCCCC}\textbf{40.98±16.29} & \cellcolor[HTML]{FFCCCC}\textbf{38.73} & \cellcolor[HTML]{FFCCCC}\textbf{58.87±12.36}  & \cellcolor[HTML]{FFCCCC}\textbf{65.46} & \cellcolor[HTML]{FFCCCC}\textbf{60.23±4.65}  & \cellcolor[HTML]{FFCCCC}\textbf{52.13} \\ \hline \hline
\end{tabular}}
\label{Main}
\end{table*}

\subsection{Effectiveness and Efficiency on Attacking VFGL (RQ1)}
In this section, NA$^{2}$ is conducted on five real-world datasets in two scenes, i.e, dual-client-based VFGL and multi-participant-based VFGL. Ablation and transferable experiments are also performed in this section. Besides, the proposed NA$^{2}$ is also compared with the query-based black-box attack regarding the number of queries needed. 

\begin{table*}  
\renewcommand\arraystretch{0.8}  
\caption{\textcolor{black}{Comparison of attack success rate between NA$^{2}$-FGA and Graph-Fraudster attack method.}}  
\centering  
\begin{tabular}{c|c|ccccc}  
\hline  
\hline  
\multicolumn{1}{r|}{\multirow{3}{*}{\textbf{Local Model}}} & \multirow{3}{*}{\textbf{Attack Method}} & \multicolumn{5}{c}{\textbf{Dataset}}\\   
\cline{3-7}   
\multicolumn{1}{r|}{}&& \textbf{Cora} & \textbf{Cora\_ML} & \textbf{Citeseer} & \textbf{Pubmed} & \textbf{ogbnn-arxiv} \\   
\hline  
\multirow{2}{*}{\textbf{GCN}} & \textbf{Graph-Fraudster} & 40\% & 28\% & 51\% & 69\% & 58\% \\  
& \textbf{NA$^{2}$-FGA} & \textbf{52\%} & \textbf{67\%} & \textbf{86\%} & \textbf{90\%} & \textbf{68\%} \\   
\hline  
\multirow{2}{*}{\textbf{SGC}} & \textbf{Graph-Fraudster} & 38\% & 46\% & 56\% & 62\% & 46\%\\  
& \textbf{NA$^{2}$-FGA} & \textbf{46\%} & \textbf{81\%} & \textbf{76\%} & \textbf{77\%} & \textbf{68\%} \\   
\hline  
\multirow{2}{*}{\textbf{GCNII}} & \textbf{Graph-Fraudster} & 39\% & 56\% & 41\% & 58\% & 47\% \\  
& \textbf{NA$^{2}$-FGA} & \textbf{62\%} & \textbf{73\%} & \textbf{69\%} & \textbf{88\%} & \textbf{76\%} \\   
\hline\hline  
\end{tabular}  
\label{Fraudster}  
\end{table*}

\subsubsection{Attack on Dual-client-based VFGL}
\begin{figure}
  \centering 
  \includegraphics[width=3.5 in]{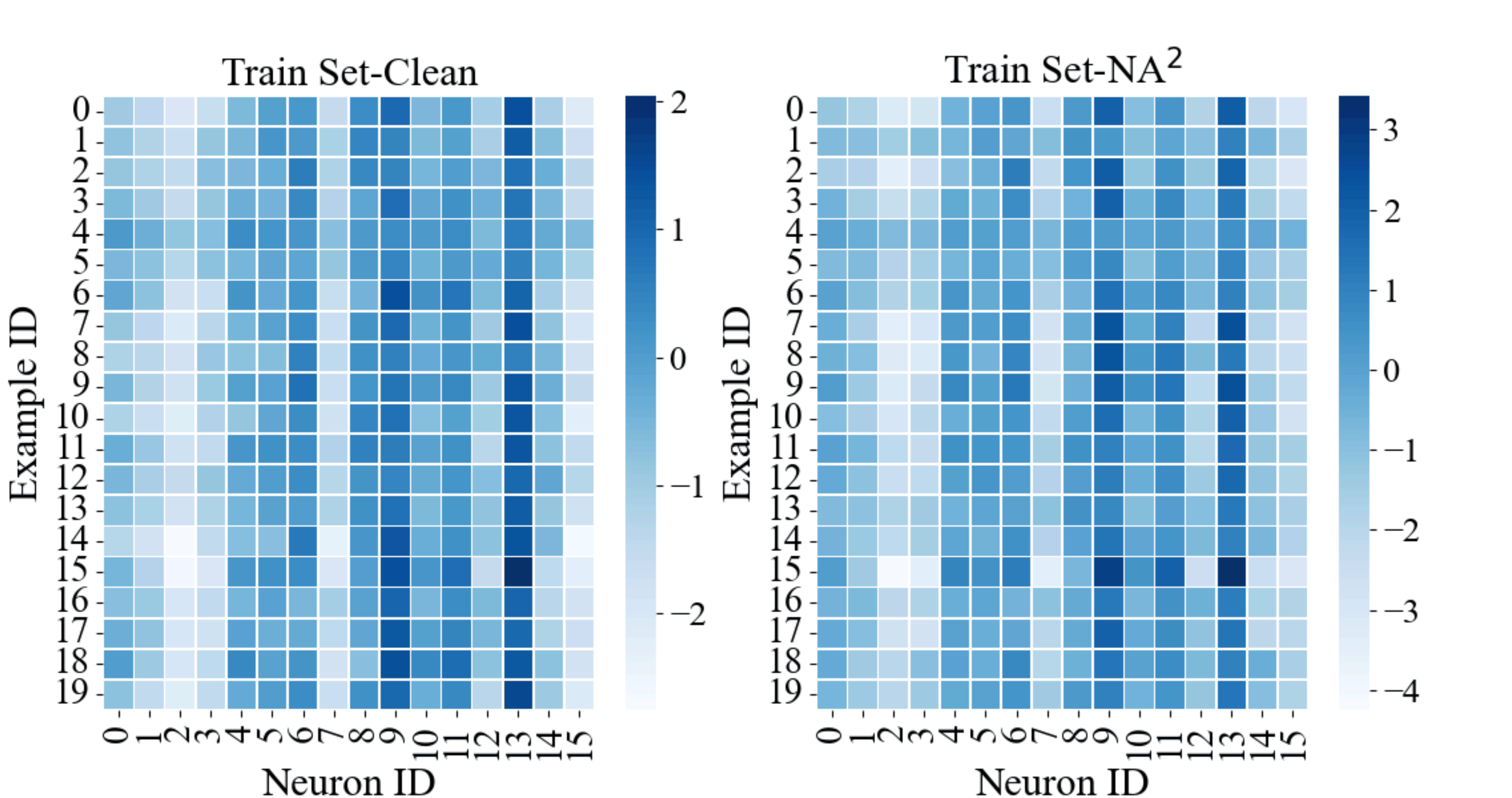}
  \caption{The effect of NA$^{2}$ on the neurons' activation of the training examples.}
  \label{fig:neuron_clean} 
\end{figure}

\begin{figure}
  \centering 
  \includegraphics[width=3.5 in]{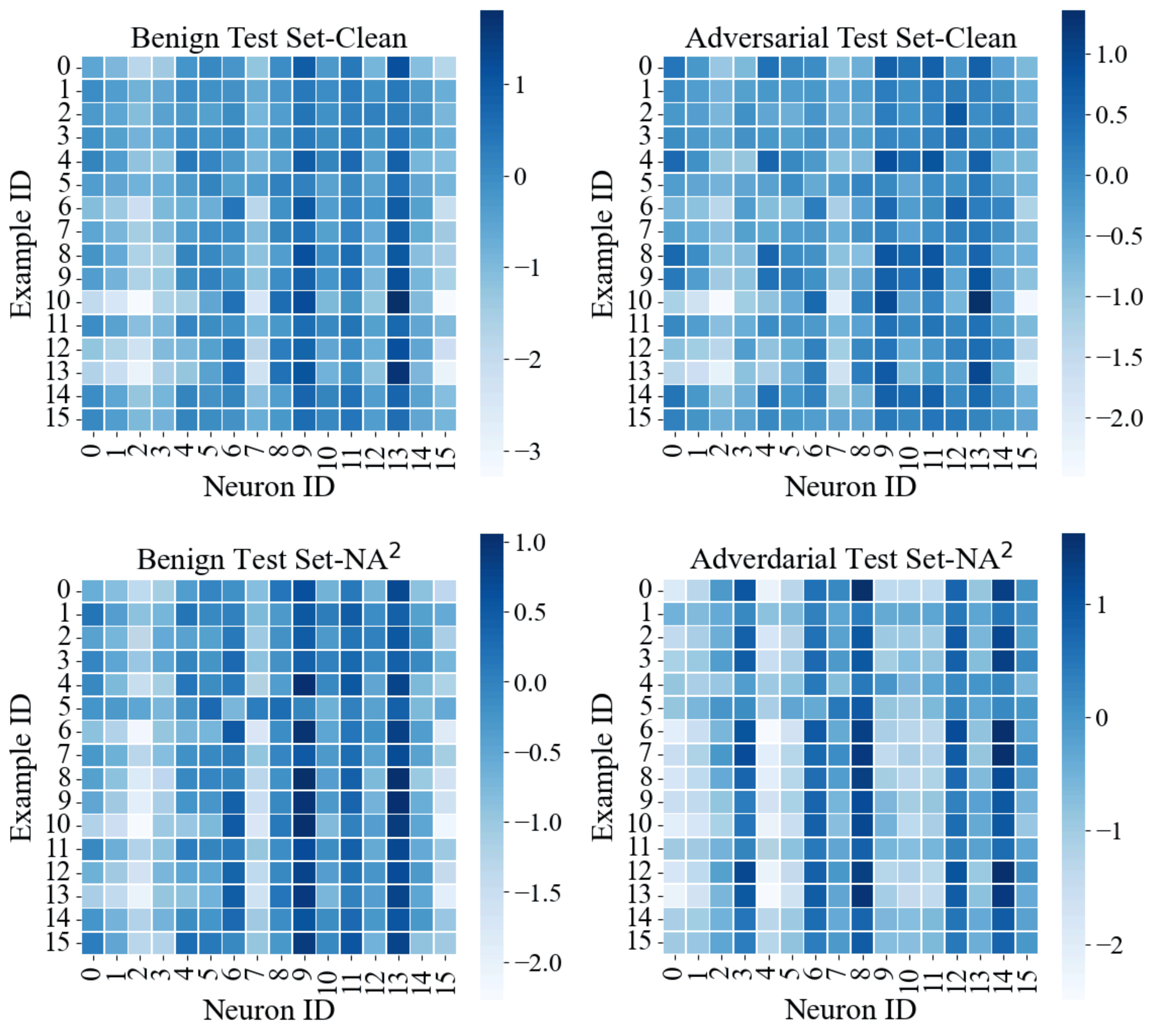}
  \caption{The effect of NA$^{2}$ on the neurons' activation of the benign examples and the adversarial examples.}
  \label{fig:neuron_ad} 
\end{figure}

As described above, in this scene, the node features are divided into two parts randomly. To verify the generality of NA$^2$, we conduct the experiments 5 times and report the ASR with standard deviation. The improvement ratio of ASR ($Impv=(\frac{ASR_{after}}{ASR_{before}}-1)\times 100\%$) is also reported, where $ASR_{before}$ is the ASR without any data manipulation while $ASR_{after}$ is the ASR using NA$^2$ or baselines. The results are shown in TABLE~\ref{Main} and TABLE~\ref{Fraudster}, and the best attack performance is highlighted in bold. Besides, we visualize the neuron's output (Fig.~\ref{fig:neuron_clean} and~\ref{fig:neuron_ad}) to illustrate the effect of NA$^2$ on the local GNN. Some observations are concluded in this experiment.

\begin{figure*}
  \centering 
  \includegraphics[width=7 in]{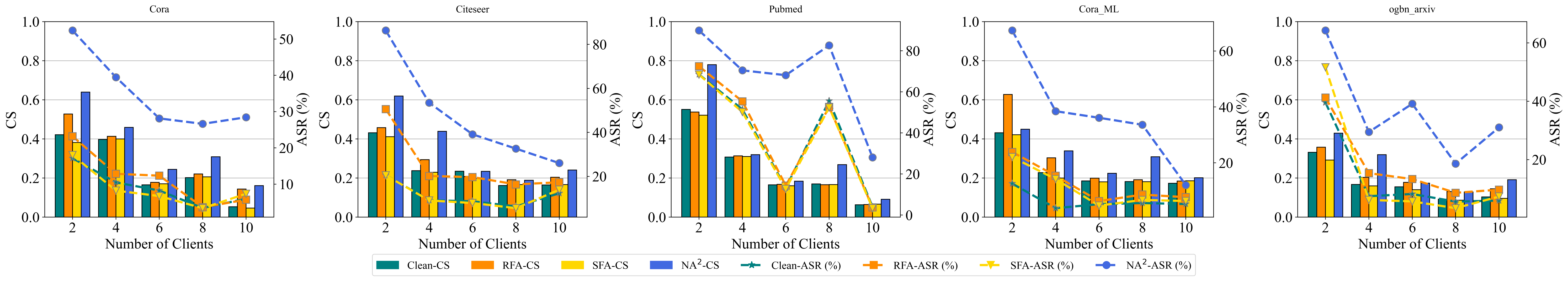}
  \caption{\textcolor{black}{ASR and CS of NA$^{2}$-SGA with different number of clients in GCN-based VFGL.}}
  \label{fig:multi} 
\end{figure*}

\emph{1) NA$^2$ can significantly improve the performance of the centralized adversarial attacks and achieves SOTA attack performance compared with baselines.} First, analyze the data in TABLE~\ref{Main}. Take FGA on GCN-based VFGL as an example, the Impv of the NA$^2$ has reached more than 200\% on Cora, Cora\_ML and Citeseer, and even achieved a 435.50\%  attack performance improvement on Cora\_ML. Under the same settings, RFA and SFA can only achieve less than 106.69\% attack performance improvement. As for Pubmed, the reason why Impv is not obvious is that the attacks have already achieved a high ASR without using any data manipulation method, but NA$^{2}$ still outperforms the baselines. Furthermore, we also note that RFA and SFA may impair attack performance due to their randomness(e.g., attack on GCN-based VFGL using GradArgmax). \textcolor{black}{For the ogbn-arxiv dataset, the attack performance of NA$^2$ decreases in comparison to the other four small datasets, but at least it improves the success rate of the attack by 27.8\% and still outperforms the baselines. The reason for the decrease in Impv value is that we believe the four attacks we selected, including FGA, are not well-suited for attacking large graph datasets. Therefore, our NA$^2$ cannot significantly improve their attack success rate.}

\textcolor{black}{Then by analyzing the results from TABLE~\ref{Fraudster}, we observe that NA$^2$-FGA consistently outperforms Graph-Fraudster across all models and datasets, achieving an average attack success rate improvement of 39.46\%. The most significant improvement is observed on the Cora\_ML dataset, with an increase of 70.00\%. We speculate the primary reasons is NA$^2$'s manipulating local training data through the steps of ``locating significant neurons path'' and ``selecting and modifying target features'' to increase the contribution of the malicious client. This ensures  the server model relies more heavily on the malicious client's data, resulting in shadow model that more accurately reflects the server model's behavior and decision boundaries. The accurate shadow model enables NA$^2$ to generate adversarial examples that are highly effective in misleading the server model, resulting in a higher attack success rate compared to Graph-Fraudster.}

\emph{2) Before and after using NA$^{2}$, the activation of neurons to training examples remains similar.} We visualized the output of neurons in the last layer of the local GNN for training examples of the same class. As shown in Fig.~\ref{fig:neuron_clean}, it can be seen that in the case of normal training, the important neurons for these examples are \#9 and \#13, and the same after using NA$^{2}$. It indicates that NA$^{2}$ does not introduce large differences on the training set, and it follows the original training process.

\emph{3) Neurons' activation to adversarial examples is changed significantly after using NA$^{2}$.} We visualize some nodes that cannot be attacked successfully under clean conditions (i.e., without data manipulation) but succeed under NA$^{2}$. In Fig.~\ref{fig:neuron_ad}, for clean conditions, the perturbations at these nodes are not enough to change the activation of neurons, which leads to the failure of the attack. As for NA$^{2}$, compared with the clean case, the distribution of neurons' activation to benign examples is more concentrated in \#9 and \#13, which is similar to the case of the training set (Fig.~\ref{fig:neuron_clean}). For adversarial examples, the distribution has changed significantly, and it is no longer \#9 and \#13. It explains why these nodes can be attacked successfully using NA$^{2}$ from the perspective of neurons.

The above experiments suggest VFGL is vulnerable to adversarial attacks as well as the centralized GNNs. Besides, the centralized attacks transferable on VFGL are not as powerful as in the centralized scene. But the threat of these attacks can be improved significantly by adopting NA$^{2}$. Then, the sensitive neurons' identification shows that the activation shift of neurons is an important factor in attack success.

\begin{figure*}
  \centering 
  \includegraphics[width=5 in]{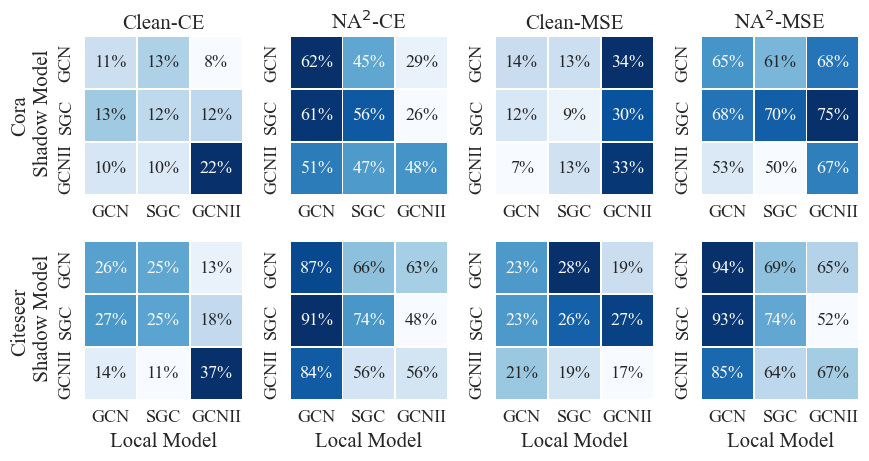}
  \caption{Transferability and ablation of NA$^2$. We present the attack performance (ASR) of nine combinations of the local GNN and the shadow model.}
  \label{fig:transfer} 
\end{figure*}
    
\subsubsection{Attack on Multi-client-based VFGL}
\textcolor{black}{Considering that a 50\% malicious client ratio is not realistic in practical scenarios,} a more realistic scenario is considered to evaluate the threat of the proposed NA$^{2}$, i.e., the multi-client-based VFGL. The number of the clients is set in \{2, 4, 6, 8, 10\}. The node features are split for each client and the structure of the graph is contained as well as the dual-client-based VFGL. The CS of the malicious client and the ASR of NA$^{2}$-SGA are reported in Fig.~\ref{fig:multi}.

In general, \textcolor{black}{similar results were presented on both large and small datasets is the} ASR of all attacks decreases as the number of clients increases, since the impact of adversarial perturbations by the malicious client will gradually decrease. In other words, when the number of clients is large enough, the perturbations of the malicious client will not be enough to confuse the server model. Therefore, improving the contribution of the malicious client to the server model, which makes the server model more dependent on the malicious client's data, is a feasible measure to improve the attack's capability in the case of multiple clients. It can be seen that NA$^{2}$ improves the CS of the malicious client and outperforms baselines in general, corresponding to a higher ASR. It is worth noting that in the case of two clients on Cora\_ML, although RFA gains a higher CS, the performance of NA$^{2}$-SGA is still better than RFA-SGA. We believe that it is because the shadow model can be better trained with the data modified by NA$^{2}$ and provides better guidance for attacks. In addition, it can be observed that in the case of eight clients on Pubmed, the performance is better than the case of six clients, which benefits from the higher CS.

In summary, in the multiple clients case, the ASR of attackers shows a downward trend with the increasing number of clients. It is caused by the declining influence of the malicious client. Compared with RFA and SFA, the effect of NA$^{2}$ on the improvement of CS is more obvious, so it can improve ASR more significantly. 


\subsubsection{Transferability and Ablation of NA$^{2}$}
The transferability and ablation of NA$^{2}$ are evaluated in this subsection. There are nine combinations (i.e., three local models and three shadow models in our experimental configuration) for a given dataset. Besides, to verify the effectiveness of the MSE, the cross entropy loss (CE) calculated from the hard labels of the training set is adopted as a comparison. The experiments are conducted on the Cora dataset and Citeseer dataset. FGA is applied as the attack generator and ASR is used as the metric. The results are shown in Fig.~\ref{fig:transfer}.

\textbf{Transferability:} Take NA$^{2}$-MSE as an example, for GCN and SGC, NA$^{2}$ has a strong transfer attack ability, that is, the shadow model established by GCN/SGC can effectively attack SGC/GCN, due to their similar structure. For GCNII, the transfer attack ability drops subtly, which can be caused by the large structural difference between GCNII and GCN (SGC) and it is difficult to establish a fully consistent shadow model. Compared with the Clean-MSE, in general, NA$^{2}$-MSE demonstrates stronger transferability (at least 50\% ASR is exhibited on both datasets, while this is only 34\% at best under Clean-MSE conditions). Similar conclusions can also be obtained in NA$^{2}$-CE.

\textbf{Ablation:} The proposed NA$^{2}$ consists of two important components, i.e., the data manipulation and the MSE-based shadow model construction. In the ablation experiments, we use these two components as ablation objects. 

Take the most general case (i.e., no data manipulation and CE is adopted as the loss function of the shadow model) as the base. Some observations are summarized as follows.

\emph{1) The data manipulation enhances the threat of the attack.} As shown in Fig.~\ref{fig:transfer}, when data manipulation and CE are applied, the ASR is greatly improved in any combination. It is improved from 8\% to 29\% even in the combination of GCN and GCNII with different structures. We believe this is due to the increased contribution of malicious parties caused by data manipulation. As a result, the perturbation has a greater impact on the server model.

\begin{table*}
\caption{Comparison of attack success rate and the average number of queries between NA$^{2}$-FGA and query-based black-box attack method. There are 100 examples randomly selected for the evaluation.}
\centering
\footnotesize
\resizebox{0.7\linewidth}{!}{
\begin{tabular}{c|c|cccccccc}
\hline
\hline
\multicolumn{1}{r|}{\multirow{3}{*}{\textbf{Local Model}}} & \multirow{3}{*}{\textbf{Attack Method}} & \multicolumn{8}{c}{\textbf{Dataset}}\\ 
\cline{3-10} 
\multicolumn{1}{r|}{}&& \multicolumn{2}{c}{\textbf{Cora}}  & \multicolumn{2}{c}{\textbf{Cora\_ML}} & \multicolumn{2}{c}{\textbf{Citeseer}} & \multicolumn{2}{c}{\textbf{Pubmed}}        
\\ \cline{3-10} 
\multicolumn{1}{r|}{} && \textbf{ASR}  & \textbf{AQ} & \textbf{ASR}    & \textbf{AQ}  & \textbf{ASR}    & \textbf{AQ}  & \textbf{ASR} & \textbf{AQ} \\ \hline
\multirow{3}{*}{\textbf{GCN}}& \textbf{GeneticAlg} & 9\% & 182.38& 3\%& 194.06& 4\%& 188.07   & 8\%& 184.49\\
& \textbf{NA$^{2}$-FGA}& \textbf{51\%}&\textbf{1.00}&\textbf{72\%}& \textbf{1.00}& \textbf{82\%}   & \textbf{1.00}  &  \textbf{87\%}& \textbf{1.00} \\ \hline
\multirow{3}{*}{\textbf{SGC}} & \textbf{GeneticAlg}& 10\%& 180.38& 6\%& 184.23 & 2\%& 198.78 & 5\%&190.21 \\
& \textbf{NA$^{2}$-FGA}& \textbf{45\%} & \textbf{1.00}& \textbf{37\%} & \textbf{1.00} & \textbf{68\%}  & \textbf{1.00} & \textbf{74\%}& \textbf{1.00}\\ \hline
\multirow{3}{*}{\textbf{GCNII}}& \textbf{GeneticAlg}& 13\%& 174.86  & 4\% & 192.31 & 10\%    & 180.45 & 10\%&180.52\\
& \textbf{NA$^{2}$-FGA}& \textbf{45\%} & \textbf{1.00}  & \textbf{73\%}   & \textbf{1.00}          & \textbf{51\%}   & \textbf{1.00} & \textbf{88\%}& \textbf{1.00}\\ 
\hline\hline
\end{tabular}
}
\label{query}
\end{table*}
    
\begin{figure*}
  \centering 
  \includegraphics[width=5.9 in]{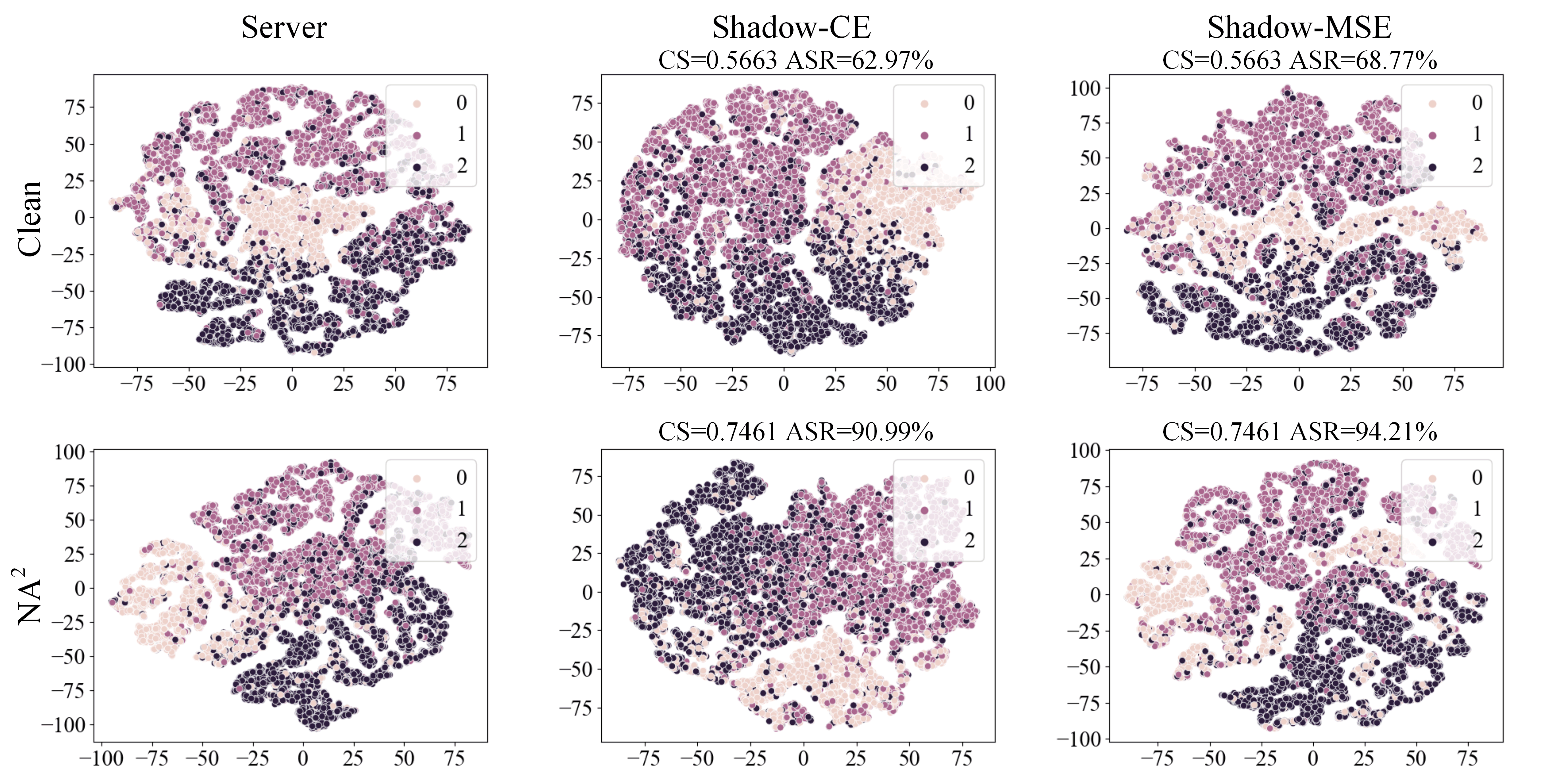}
  \caption{The embeddings which obtained from the server model and the shadow model in the GCN-based VFGL on the Pubmed dataset. They are projected into a 2-dimensional space using t-SNE. The CS and ASR of SGA are reported at the same time.}
  \label{fig:tsne} 
\end{figure*}

\emph{2) The MSE-based shadow model makes the attack more transferable.} While the MSE is adopted without data manipulation, the performance of the attack is slightly improved, and the phenomenon is more obvious in the transfer attack. For example, on the Cora dataset, using GCN as the shadow model to attack, the ASR increases from 8\% to 34\%. In addition, the improvement is more significant between models with large structural differences, because MSE can effectively utilize the behavioral information leaked by probabilities, which is more conducive to the establishment of the shadow model. To better illustrate it, taking the Pubmed dataset as an example, we project the embeddings extracted from the server model and the shadow model by using t-SNE, which is shown as Fig.~\ref{fig:tsne}. It can be observed from the distribution of t-SNE that the distribution using CE is significantly different from the target distribution (Server). It shows that CE only pays attention to the performance but ignores the behavior of the server model. The t-SNE distribution extracted from the shadow model built with NA$^{2}$ using MSE is most similar to the server model with NA$^{2}$. Besides, the CS and ASR (SGA) of the shadow model with NA$^{2}$-MSE show the best attack performance among all the combinations.

\textcolor{black}{Many existing works~\cite{song2024two,jin2021catchbackdoor} use t-SNE to visualize the embeddings in order to explain the effectiveness of the attack methods. Therefore, inspired by this, we use t-SNE for visualization. For NA$^{2}$, the effectiveness of it is closely related to the construction of the shadow model. The higher the similarity between the shadow model and the server's behavior, the more accurate the information about the server that the shadow model provides to the attacker, and NA$^{2}$ performs better. That’s why we use t-SNE to visualize the embeddings of the shadow model and server.}

In summary, combining the advantages of the above two components (i.e., the data manipulation and the MSE-based shadow model), NA$^{2}$-MSE achieves SOTA performance in both ASR and transfer attack ability. 

\subsubsection{Efficiency of Attacking VFGL}
In order to verify the query efficiency of NA$^{2}$, we compare NA$^{2}$-FGA with a query-based black-box attack method, namely GeneticAlg. GeneticAlg is a genetic algorithm-based attack method. The population of GeneticAlg is set to 20, and the number of iterations is set to 10. Thus, the maximum number of queries for a target node is 200. If the attack fails, the number of queries is fixed to 200. TABLE~\ref{query} reports the ASR and AQ of NA$^{2}$-FGA and GeneticAlg on four datasets and three local GNNs.

Since NA$^{2}$-FGA only needs to initiate one request to the server model to get the probabilities, the number of queries for NA$^{2}$-FGA is always fixed to 1. As for GeneticAlg, more queries are necessary to optimize the perturbations. Therefore, GeneticAlg has far more queries than NA$^{2}$-FGA. Besides, in each setting, the ASR of NA$^{2}$-FGA far exceeds that of GeneticAlg, which shows the superiority of NA$^{2}$-FGA in attack performance and query efficiency.

\textcolor{black}{In federated learning, stragglers refer to the nodes that take longer to complete the computation tasks compared to other participating nodes. This lag may slow down the training progress of the entire system because the training process usually needs to wait for the computation results from all nodes before it can proceed. Stragglers may appear due to factors such as differences in hardware performance, network delays, and uneven computation loads. However, stragglers will not affect the number of queries for NA$^{2}$. The main reason is that NA$^{2}$'s queriy occur after the server model has aggregated, i.e., (after all clients have uploaded their local node embeddings). Therefore, NA$^{2}$ can query normally even after stragglers appear, and the number of queries remains 1, with only a delay in the query time.}

\begin{table}[htbp]
\centering
\caption{\textcolor{black}{Significance analysis among CS (C), MSE (M) and ASR (A).}}
\resizebox{0.9\linewidth}{!}{\LARGE
\begin{tabular}{c|ccc|ccc}
\hline
\hline
\multirow{3}{*}{\textbf{Percentage}}&\multicolumn{3}{c|}{\textbf{p-value}}&\multicolumn{3}{c}{\textbf{Correlation Coefficient}}\\ \cline{2-7}&\textbf{C-M}&\textbf{M-A}&\textbf{C-A}&\textbf{C-M}&\textbf{M-A}&\textbf{C-A}\\ \hline
\textbf{50\% }&2.2E-11&2.8E-11&2.2E-07&-0.93&-0.75&0.91\\ \hline
\textbf{25\% }&2.30E-07&2.60E-07&8.5E-08&-0.58&-0.65&0.87\\ \hline
\textbf{12.5\%}&3.4E-04&4.9E-04&4.6E-04&-0.35&-0.45&0.47\\ \hline \hline
\end{tabular}}
\label{p-value}
\end{table}

\subsection{Relationship between the ASR and the Contribution (RQ2)}
In this subsection, we perform the significance analysis for the ASR, the MSE and the CS of the malicious client. The correlation coefficient and p-value are adopted as the metrics for the analysis. \textcolor{black}{We set the number of clients to \{2, 4, 8\} and did three sets of experiments, since there is only one malicious client in our attack method, the percentage of malicious client in the experiments is \{50\%, 25\%, 12.5\%\}. The analysis are shown in TABLE~\ref{p-value}. The analysis is based on GCN-based VFGL on the Cora dataset.} 

 The p-value indicates that there is a strong significance between CS and MSE, MSE and ASR, CS and ASR (the p-value less than 0.05 is significant, and the p-value less than 0.01 is very significant). The correlation coefficient suggests there is a high negative correlation between CS and MSE, MSE and ASR (i.e., the higher the contribution, the better the shadow model can be established with the server, and the shadow model with high similarity can better guide the adversarial attack). And there is a strong positive correlation between CS and ASR, i.e., the VFGL is more likely to be attacked by the high-contribution client.

\begin{figure}
	\centering  
	\vspace{-0.35cm}
	\hspace{-0.8cm}
	\subfigtopskip=2pt 
	\subfigbottomskip=2pt 
	\subfigcapskip=-5pt
	\subfigure[without NA$^{2}$]{
		\label{weight.sub.1}
		\includegraphics[width=0.5\linewidth]{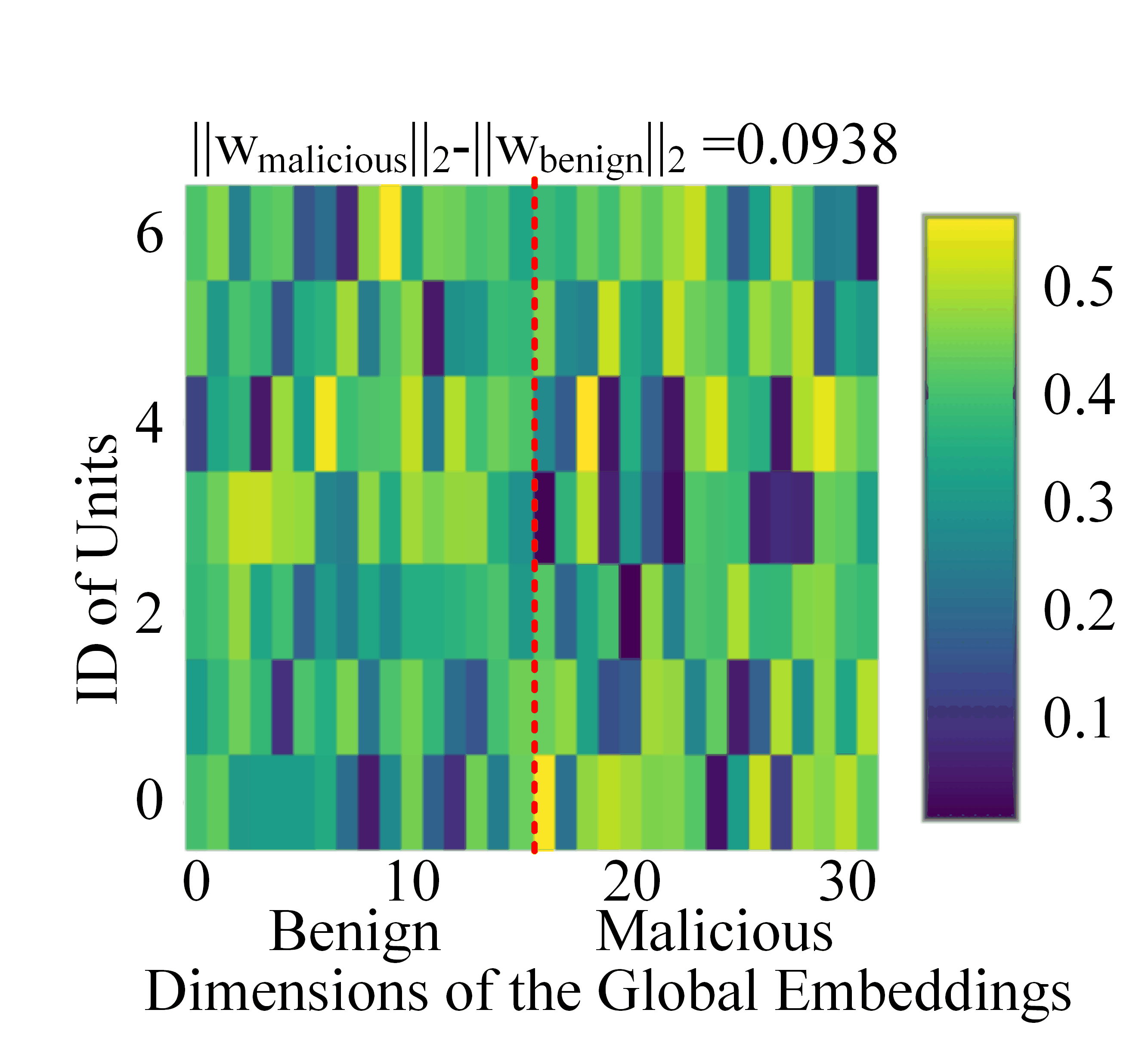}}
	\subfigure[with NA$^{2}$]{
		\label{weight.sub.2}
		\includegraphics[width=0.5\linewidth]{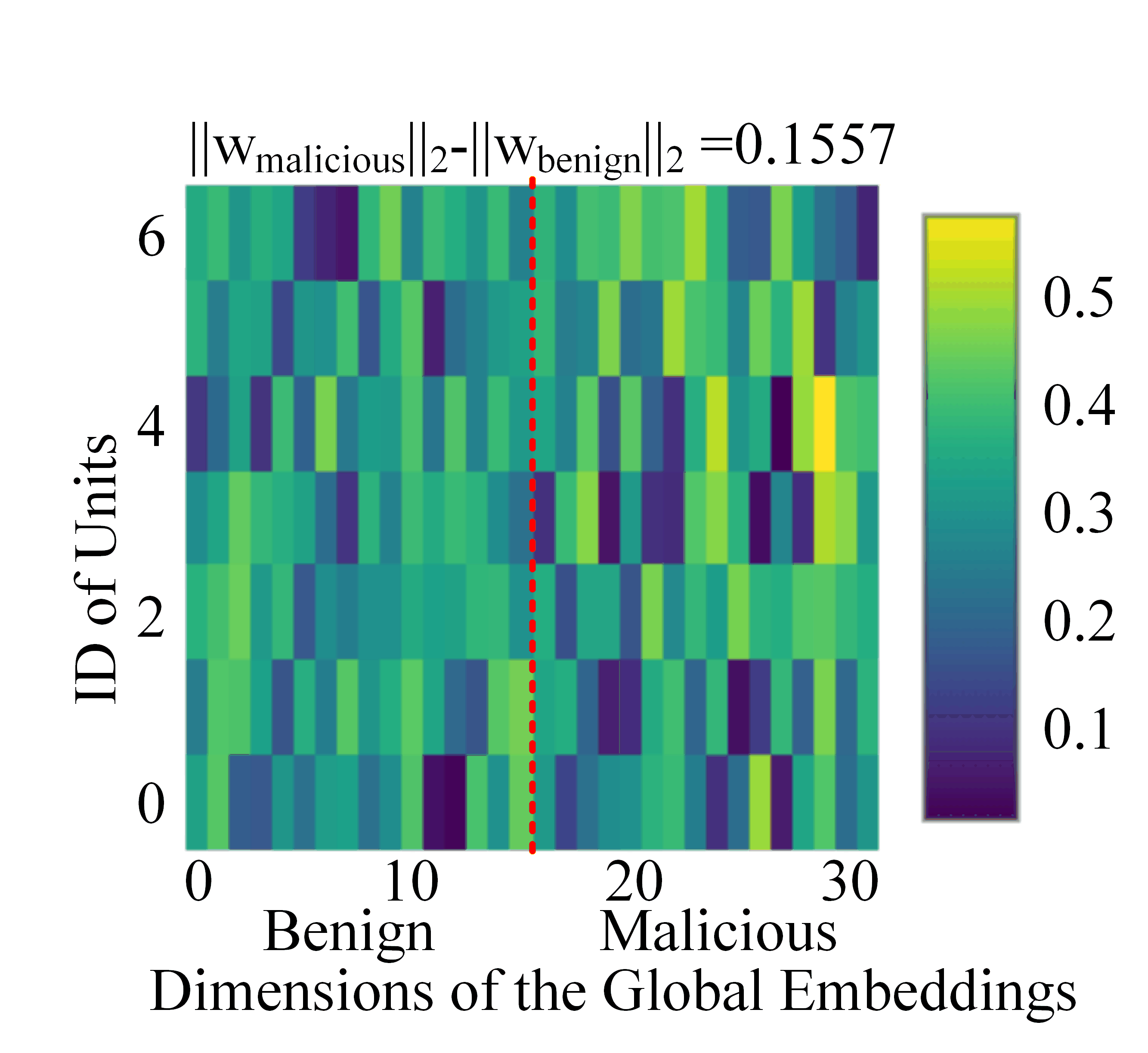}}
	\caption{Weights distribution of the server's input layer wo/w NA$^{2}$.}
	\label{fig:weight}
\end{figure}

In addition, from the perspective of the server model, we visualize the weights of the input layer of the server model in dual-client, which is shown in Fig.~\ref{fig:weight}. Since the server model cascades node embeddings from the clients, the weights of the input layer can reflect how dependent the server model is on different clients. $||w_{malicious}||_{2}-||w_{benign}||_{2}$ is used to represent the difference in the model's dependence on malicious and benign clients, where $||\cdot||_{2}$ is $L2$-norm, $w_{malicious}$ and $w_{benign}$ is the weight of the malicious client and the benign client respectively. $||w_{malicious}||_{2}-||w_{benign}||_{2}$ is improved from 0.0938 to 0.1557 after adopting NA$^2$, which indicates the server model is more dependent on the malicious client.

\subsection{The Performance of NA$^{2}$ with the Possible Defense (RQ3)}
To mitigate the performance of NA$^{2}$, differential privacy (DP)~\cite{dwork2008differential} can be adopted as a potential defense method. Unlike approaches such as homomorphic encryption, DP focuses on noise injection on the embeddings rather than encryption in transit, which will be decrypted on the server-side. Therefore, the predicted probabilities of the server model will be changed to weaken the ability of NA$^{2}$ to build the shadow model. In this work, the Laplacian mechanism is used to produce the noises with the scale of noise $\epsilon\in [0,0.5]$. The experiments are validated on Cora and Citeseer, and SGA is selected as the attack generator.

TABLE~\ref{defense} shows that with the increase of the noise scale, the attack performance of NA$^{2}$-SGA decreases slightly (within 10\%). Besides, injecting large-scale noise will significantly affect the performance of VFGL. For example, when the $\epsilon=0.5$, the accuracy of VFGL dropped by 6.03\%, 3.20\%, 7.00\% and 9.90\% on each dataset, respectively. Using DP requires a trade-off between model performance and protection capabilities. In general, DP cannot prevent NA$^{2}$.

\textcolor{black}{In addition to DP, FoolsGold~\cite{fung2018mitigating} can also be considered as a potential defense method. It measures the similarity between clients by calculating the cosine similarity of historical gradient updates and uses a historical mechanism to track updates from each client. To adapt to the VFL scenario, we require the server to be able to access the gradient information of the client's model when using FoolsGold. Notably, this change is advantageous to the defender because in the GVFL scenario, the server can only receive embeddings uploaded by the client without gradient information. We set the number of clients to \{2, 6, 10\} and conducted experiments on four datasets. In the experiments, the attack generator also chose SGA, and the evaluation indicators used were detection rate (DR) and ASR, where DR is the ratio of the number of experiments in which malicious clients were successfully detected to the total number of experiments. The results are shown in TABLE~\ref{foolsgold}.}

\textcolor{black}{The results show that the average detection rate of FoolsGold is 8.98\%, and the detection rate gradually decreases as the number of clients increases. In addition, the average ASR after defense only decreases by 9.29\%, indicating that NA$^{2}$ has good concealment. The core idea of FoolsGold is to distinguish malicious clients from honest clients by utilizing the diversity of client gradient updates. When multiple Sybil clients collaborate in an attack, they share a malicious goal, resulting in more similar gradient updates, which can be identified and weakened by FoolsGold. However, NA$^{2}$ only has one malicious client, and a single malicious client cannot form collaboration with other clients, making it difficult to trigger the detection mechanism of FoolsGold.}

\subsection{Parameter Analysis and Time Complexity Analysis of NA$^{2}$ (RQ4)}
In this subsection, we explore the sensitivity of the feature modification scale $\gamma$ and the starting epoch $\tau$ of data manipulation. In more detail, $\gamma$ is set from 0 to 0.1 with the step size of 0.01, and $\tau$ is set from 0 to 50 with the step size of 5. The experiments are carried out on the Cora dataset. In addition, the time complexity is analyzed and demonstrated with running time experiments.

\begin{table}
\renewcommand\arraystretch{0.9}
\centering
\caption{The performance of NA$^{2}$-SGA against the possible defense (DP) on GCN-based VFGL.}
\resizebox{0.9\linewidth}{!}{\LARGE
\begin{tabular}{c|c|cccccc}
\hline
\hline
\multirow{2}{*}{\textbf{Dataset}}  & \multirow{2}{*}{\textbf{Metric}} & \multicolumn{6}{c}{\textbf{$\epsilon$}}                                     \\ \cline{3-8} 
                          &                         & \textbf{0}        & \textbf{0.1 }     & \textbf{0.2 }     & \textbf{0.3}      & \textbf{0.4}      & \textbf{0.5}      \\ \hline
\multirow{2}{*}{\textbf{Cora}}     & \textbf{ACC (\%)}                 & 76.70 & 76.80    & 74.70 & 74.60 & 72.90 & 70.80    \\
                          & \textbf{ASR (\%)}                 & 59.58 & 49.74 & 51.67   & 54.56 & 56.79 & 55.65 \\ \hline
\multirow{2}{*}{\textbf{Cora\_ML}}     & \textbf{ACC (\%)}                 & 77.90 & 77.00    & 74.60 & 74.30 & 74.10 & 74.70    \\
                          & \textbf{ASR (\%)}                 & 51.35 & 49.74 & 45.04   & 47.78 & 41.57 & 46.85 \\ \hline
\multirow{2}{*}{\textbf{Citeseer}} & \textbf{ACC (\%)}                & 66.80    & 67.90    & 66.90    & 62.40    & 64.40    & 59.80    \\
                          & \textbf{ASR (\%)}                & 85.48   & 80.27   & 77.43   & 87.98   & 83.23   & 77.26   \\ \hline 
\multirow{2}{*}{\textbf{Pubmed}}     & \textbf{ACC (\%)}                 & 77.70 & 63.70    & 69.90 & 69.90 & 69.10 & 67.80    \\
                          & \textbf{ASR (\%)}                 & 94.21 & 90.58 & 84.98   & 88.13 & 90.45 &90.71 \\ \hline \hline
\end{tabular}}
\label{defense}
\end{table}

\begin{table}
\renewcommand\arraystretch{0.9}
\centering
\caption{\textcolor{black}{The performance of NA$^{2}$-SGA against the possible defense ( FoolsGold) on GCN-based VFGL ( The data in the parentheses indicates the ASR  when no defense measures are taken).}}
\resizebox{0.9\linewidth}{!}{\Large
\begin{tabular}{c|c|ccc}
\hline
\hline
\multirow{2}{*}{\textbf{Dataset}}  & \multirow{2}{*}{\textbf{Metric}} & \multicolumn{3}{c}{\textbf{Number of Client}}                                      \\ \cline{3-5} 
                          &                         & \textbf{2}        & \textbf{6 }     & \textbf{10 }                      \\ \hline
\multirow{2}{*}{\textbf{Cora}}     & \textbf{DR (\%)}                 & 12.76     & 6.43  & 3.11    \\
                          & \textbf{ASR (\%)}                 & 49.56(52.37)  & 25.33(28.21)    & 27.45(28.51)  \\ \hline
\multirow{2}{*}{\textbf{Cora\_ML}}     & \textbf{DR (\%)}                 & 16.18     & 9.61  & 4.67     \\
                          & \textbf{ASR (\%)}                 & 60.75(67.34)   & 31.44(36.12)    & 11.86(12.45)   \\ \hline
\multirow{2}{*}{\textbf{Citeseer}} & \textbf{DR (\%)}                & 13.84       & 7.29       & 3.31        \\
                          & \textbf{ASR (\%)}                & 78.96(86.34)       & 35.33(39.35)     & 26.54(26.04)      \\ \hline 
\multirow{2}{*}{\textbf{Pubmed}}     & \textbf{DR (\%)}                 & 17.52     & 8.57  & 4.47     \\
                          & \textbf{ASR (\%)}                 & 74.58(89.91)   & 62.58(68.64)   & 26.97(28.42)  \\ \hline \hline
\end{tabular}}
\label{foolsgold}
\end{table}

\begin{figure}[!t]
	\centering  
	\vspace{-0.1cm}
	\hspace{-0.8cm}
	\subfigure[scale of features modification]{
		\label{parameters.sub.1}
		\includegraphics[width=0.515\linewidth]{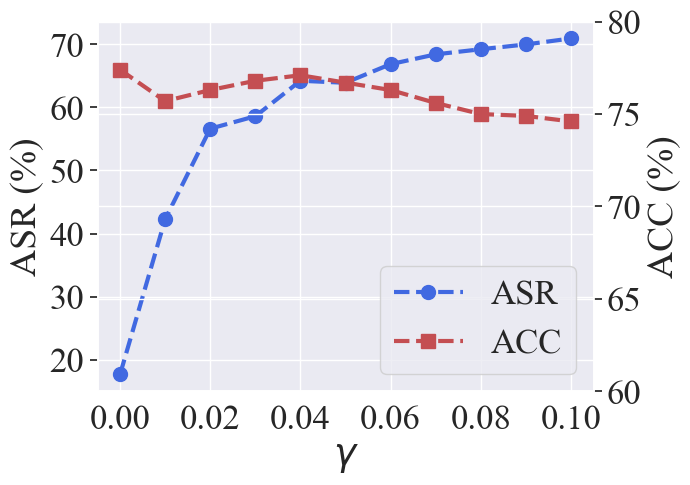}}
	\hspace{-0.2cm}
	\subfigure[epoch to start NA$^{2}$]{
		\label{parameters.sub.2}
		\includegraphics[width=0.53\linewidth]{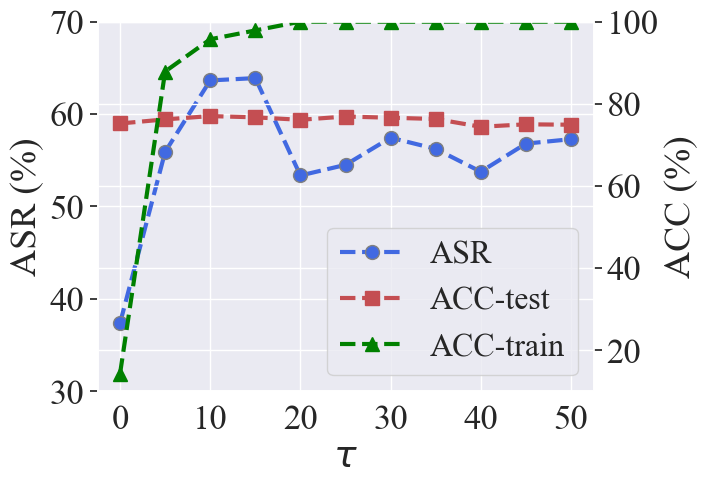}}
	\hspace{-0.5cm}
	\caption{Performance of NA$^{2}$-FGA on the Cora dataset with different scale of features modification and different epochs to start features modification on GCN-based VFGL.}
	\label{fig:parameters}
\end{figure}

\begin{figure}[!t]
	\centering  
	\vspace{-0.1cm}
	\hspace{-0.8cm}
	\subfigure[time for data manipulation]{
		\label{time.sub.1}
		\includegraphics[width=0.51\linewidth]{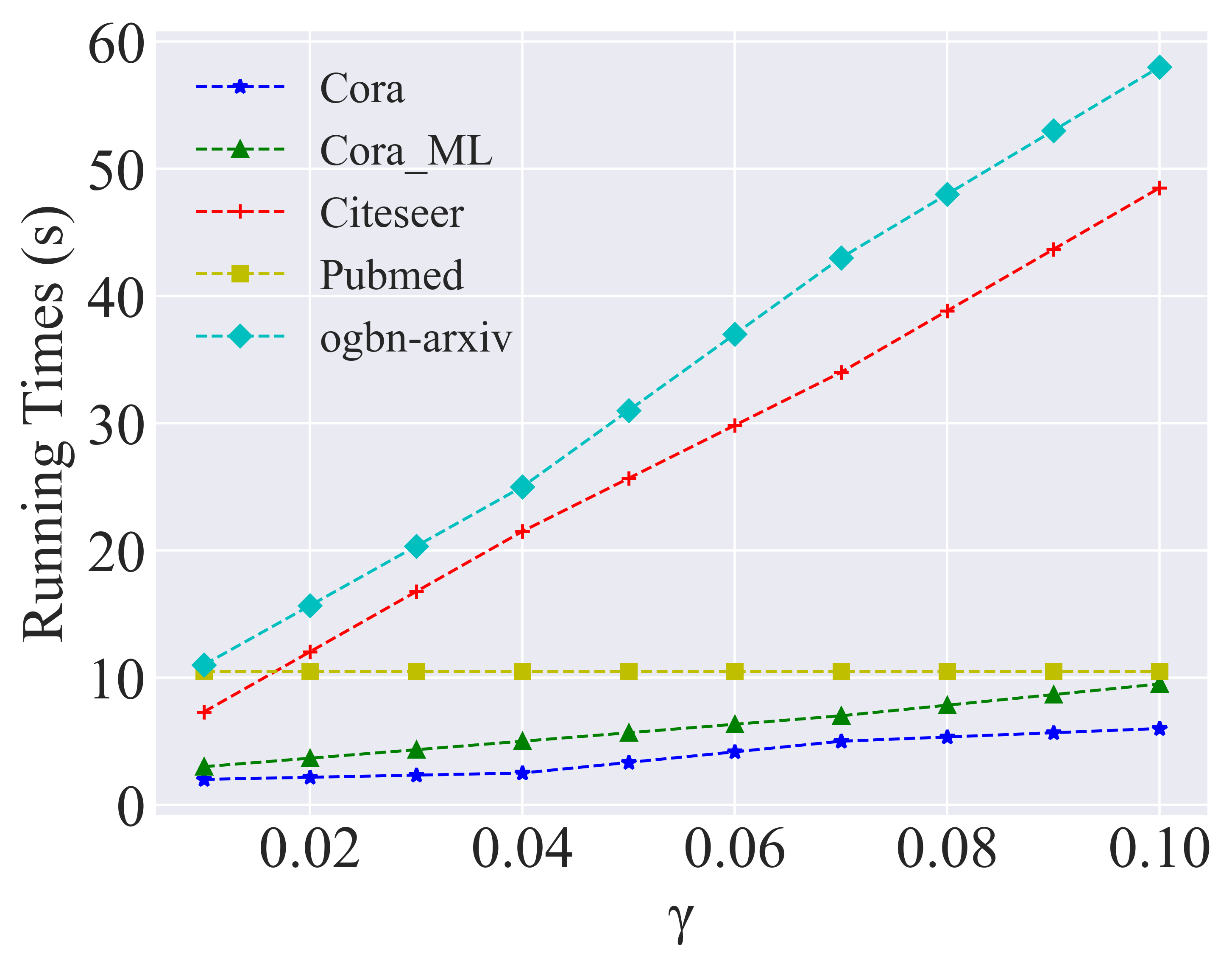}}
	\hspace{-0.2cm}
	\subfigure[time for each stage of NA$^{2}$-SGA]{
		\label{time.sub.2}
		\includegraphics[width=0.53\linewidth]{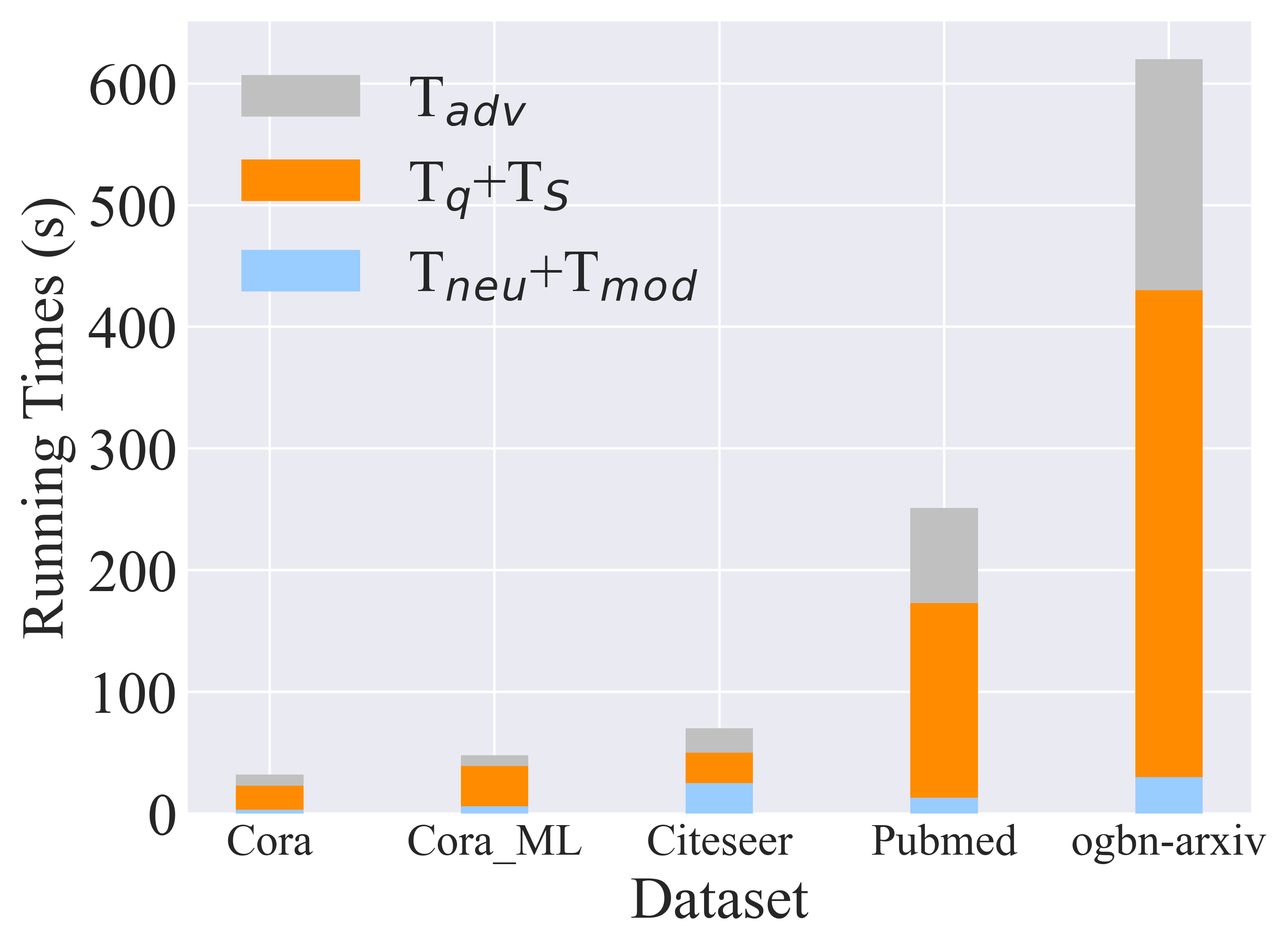}}
	\hspace{-0.5cm}
	\caption{\textcolor{black}{Running time of NA$^{2}$.}}
	\label{fig:time}
\end{figure}

\textbf{Parameter analysis:} The performance of NA$^{2}$-FGA is shown in Fig.~\ref{fig:parameters} with the change of the parameters. In Fig.~\ref{parameters.sub.1}, it can be observed that the ASR increases gradually with the increase of $\gamma$, and finally tends to be stable. However, it also can be noticed that an excessively large $\gamma$ also affects the benign accuracy of VFGL. For $\tau$, as shown in Fig.~\ref{parameters.sub.2}, the ASR rises rapidly in the early stage of training (before $\tau=15$), which may be caused by the uncertainty of important neurons in the early stage. Then, the ASR fluctuates in a small range (within 10\%), because the server model has already achieved high accuracy on the training set. In summary, for $\gamma$, we suggest using a moderate value (e.g., 0.05) to trade off the attack performance and the VFGL's performance. As for $\tau$, data manipulation at an early training stage may affect the attack performance while the model has not started converging.

\textbf{Time complexity analysis:} The time cost of NA$^{2}$ mainly comes from five parts, including the time cost for locating the neurons ($T_{neu}$), the time cost ($T_{mod}$) to modify the features of the candidate nodes, one querying time cost ($T_{q}$), the shadow model's training time cost ($T_{\Tilde{S}}$), and the time cost for the attacker to generate the adversarial perturbations ($T_{adv}$). Therefore, the time complexity of NA$^{2}$ is
\begin{equation}
\small
    {\mathcal O}(T_{neu})+{\mathcal O}(T_{mod})+{\mathcal O}(T_{q})+{\mathcal O}(T_{\Tilde{S}})+{\mathcal O}(T_{adv})\sim{\mathcal O}(N),
\end{equation}
where ${\mathcal O}(T_{neu})$ depends on the number of the testing examples and the scale of the graph. ${\mathcal O}(T_{mod})$ depends on the budget of the modified features, and ${\mathcal O}(T_{q})$ is the complexity of the one-querying time. ${\mathcal O}(T_{\Tilde{S}})$ is the complexity of the training time of the shadow model.  ${\mathcal O}(T_{adv})$ depends on the adversarial attack methods. \textcolor{black}{Different adversarial scenarios may require different perturbation generators, and the time complexity of different perturbation generators may vary. However, the design of the NA$^{2}$  allows it to adapt to various perturbation generators, so we can maintain the linearity of $T_{adv}$ by choosing an appropriate perturbation generator.} Thus, according to all the above steps, ${\mathcal O}(N)$ indicates that the time complexity of NA$^{2}$ is linear.

Further, the running time of NA$^{2}$ is tested. As analyzed above, since the shadow model training time cost ($T_{\Tilde{S}}$) and the attack generator's attack time cost ($T_{adv}$) are linear. Compared to other steps, the query time cost ($T_{q}$) is negligibly short. Thus, only the running time of data modification ($T_{neu}+T_{mod}$) is tested, and the results are shown in Fig.~\ref{time.sub.1}. It can be seen that the running time is linear with the feature modification scale $\gamma$, which is consistent with the analysis of time complexity. It takes more time on the Citeseer dataset because Citeseer has more nodes than Cora and Cora\_ML, the number of features is much larger than Pubmed. Besides, the running time for each stage of NA$^{2}$ is shown in Fig.~\ref{time.sub.2}. The SGA is adopted as the perturbation generator and 100 examples are tested for running time. \textcolor{black}{Obviously, the time consumption for $T_{neu}+T_{mod}$ is the least. We notice that the time cost for the data manipulation on the Citeseer dataset occupies a larger proportion of the total time cost, which is consistent with Fig.~\ref{time.sub.1}. On the Pubmed dataset, it takes more time to train the shadow model due to its large scale. Furthermore the time complexity of NA$^{2}$ remains linear when using a large dataset ogbn-arxiv. The step of “locating the neuron” involves traversing all layers of the local model for each node according to certain rules. $T_{neu}$ of each node in this step is similar, as the complexity of the graph increases, the total number of nodes increases linearly, so the final time complexity remains linear.}

\section{Discussion}
\label{discussion}
In this section, we make the discussion about the proposed NA$^{2}$, including limitation, future work and the comparison with the backdoor attacks.

\subsection{Limitation and Future Work}

\hspace{2em}\textbf{Limitation:} We test the VFGLs' accuracy on benign data before and after using NA$^{2}$, the results are reported in TABLE~\ref{clean}. The results show that the performance of VFGL after using NA$^{2}$ decreases slightly. This is understandable that NA$^2$ makes server model more dependent on the data from the malicious clients. The data from the benign clients may not be fully utilized, which further causes the decrease of the main task accuracy. 

\textbf{Future work:} In future work, we aim to improve the contribution of the malicious client on the premise of making full use of each client's data to ensure the performance of the main task.

\begin{table}
\large
\centering
\caption{The accuracy (\%) of the VGFL wo/w NA$^2$.}
\resizebox{8.9cm}{!}{
\begin{tabular}{c|cccccc}
\hline\hline
\multirow{3}{*}{\textbf{Dataset}} & \multicolumn{6}{c}{\textbf{Local Model}}                                                                                                     \\ \cline{2-7} 
                                  & \multicolumn{2}{c|}{\textbf{GCN}}                  & \multicolumn{2}{c|}{\textbf{SGC}}                  & \multicolumn{2}{c}{\textbf{GCNII}} \\  \cline{2-7}
                                  & \textbf{Clean} & \multicolumn{1}{c|}{\textbf{NA$^2$}} & \textbf{Clean} & \multicolumn{1}{c|}{\textbf{NA$^2$}} & \textbf{Clean}    & \textbf{NA$^2$}   \\ \hline
\textbf{Cora}                     & 78.26±0.56          & \multicolumn{1}{c|}{74.24±2.42}        & 78.14±0.97          & \multicolumn{1}{c|}{76.42±1.09}        & 77.18±1.84             & 74.24±2.32          \\ \hline
\textbf{Cora\_ML}                 & 81.92±1.17          & \multicolumn{1}{c|}{74.58±2.66}        & 81.78±0.65          & \multicolumn{1}{c|}{75.98±6.23}        & 79.08±1.95             & 76.06±3.52          \\ \hline
\textbf{Citeseer}                 & 65.24±1.25          & \multicolumn{1}{c|}{64.10±2.74}        & 65.24±1.84          & \multicolumn{1}{c|}{62.58±2.14}        & 59.08±2.54             & 58.18±3.72          \\ \hline
\textbf{Pubmed}                   & 77.56±0.71          & \multicolumn{1}{c|}{75.62±3.48}        & 77.70±1.28          & \multicolumn{1}{c|}{73.00±0.10}        & 74.94±1.97             & 72.76±2.90          \\ \hline \hline
\end{tabular}}
\label{clean}
\end{table}


\subsection{Comparison with the Backdoor Attacks}
It is worth noting that NA$^{2}$ is a kind of hybrid attack, which contains the data manipulation in the training process and the adversarial attack in the testing process. In order to distinguish NA$^{2}$ and existing works of backdoor attacks, in this subsection, we discuss the comparison of NA$^{2}$ with the backdoor attacks on GNNs and FL.

\textbf{Comparison with the backdoor attacks on GNNs:} Existing works on graph backdoor~\cite{xi2021graph,zhang2021backdoor} focus on injecting backdoors (triggers) into the training data and triggering attacks through the fixed triggers (e.g., fixed subgraphs). Although NA$^{2}$ manipulates the training data, its purpose is to facilitate the adversarial attacks in VGFL scenario where features of the global model are not well known to the malicious client/attacker. The adversarial attack using NA$^{2}$ can be launched using flexible perturbations generated during the inference time, however, the backdoor attack requires a fixed trigger that is embedded during the training. 

\textbf{Comparison with the backdoor attacks on FL:} Different from the centralized backdoor attack (CBA)~\cite{gu2017badnets} and the distributed backdoor attack (DBA)~\cite{xie2020dba},  NA$^{2}$ can leverage diverse adversarial perturbations to achieve attacks on any target example, instead of using fixed local triggers or distributed triggers.

\section{Conclusion}
\label{conclusion}
In this work, we proposed a query-efficient adversarial attack framework named NA$^{2}$. The proposed NA$^{2}$ uses the key neuron path to locate the key neurons and modifies the corresponding important node features to improve the contribution of the malicious client. A shadow model of the server model is established via the manipulated data and the probabilities returned by the server model. Extensive experiments prove that NA$^{2}$ can improve the performance of the existing centralized adversarial attacks against VFGL significantly, and can achieve SOTA performance. Even with the deployment of the defense mechanism, NA$^{2}$ still achieves impressive attack results. In addition, the sensitive neurons identification and visualization of t-SNE are provided to understand the effectiveness of NA$^{2}$.

\ifCLASSOPTIONcompsoc
  \section*{Acknowledgments}

\fi

This research was supported by the National Natural Science Foundation of China (Nos. 62072406 and U21B2001),
and the Zhejiang Provincial Natural Science Foundation (No. LDQ23F020001).

\ifCLASSOPTIONcaptionsoff
  \newpage
\fi



%



\bibliographystyle{IEEEtran}
\bibliography{ref.bib}

%
\vspace{-10 mm}
\begin{IEEEbiography}[{\includegraphics[width=0.95in,height=1.2in,clip, keepaspectratio]{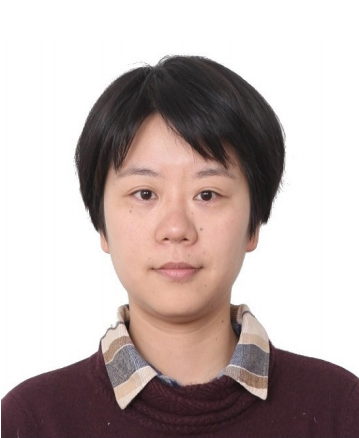}}]{Jinyin Chen} received BS and PhD degrees from Zhejiang University of Technology, Hangzhou, China, in 2004 and 2009, respectively. She studied evolutionary computing in Ashikaga Institute of Technology, Japan in 2005 and 2006. She is currently a Professor with the Zhejiang University of Technology, Hangzhou, China. Her research interests include artificial intelligence security, graph data mining and evolutionary computing
\end{IEEEbiography}

\vspace{-10 mm}
\begin{IEEEbiography}[{\includegraphics[width=0.95in,height=1.15in,clip,keepaspectratio]{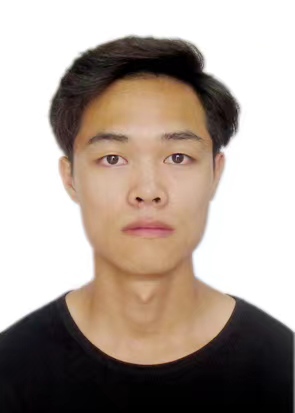}}]{Wenbo Mu} is currently pursuing the masters degree with the college of Information engineering, Zhejiang University of Technology. His research interests include graph data mining and application, and artificial intelligence.
\end{IEEEbiography}

\vspace{-10 mm}
\begin{IEEEbiography}[{\includegraphics[width=0.95in,height=1.15in,clip,keepaspectratio]{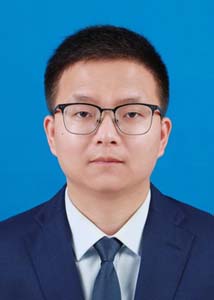}}]{Luxin Zhang} is currently pursuing the masters degree with the college of Information engineering, Zhejiang University of Technology. His research interests include graph data mining and application, and artificial intelligence.
\end{IEEEbiography}

\vspace{-10 mm}
\begin{IEEEbiography}[{\includegraphics[width=0.95in,height=1.15in,clip, keepaspectratio]{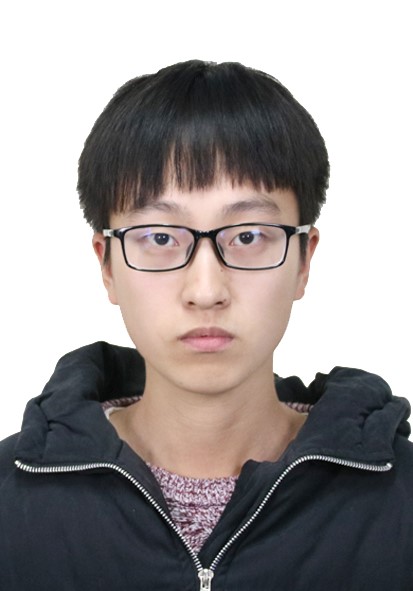}}]{Guohan Huang} is currently pursuing the master degree with the college of Information engineering, Zhejiang University of Technology. His research interests include graph data mining and applications, and artificial intelligence.
\end{IEEEbiography}

\vspace{-10 mm}
\begin{IEEEbiography}[{\includegraphics[width=0.95in,height=1.15in,clip,keepaspectratio]{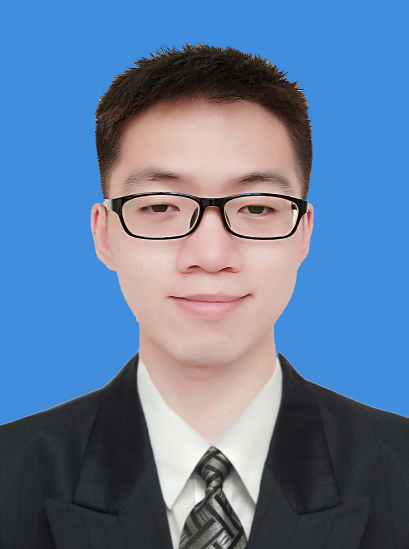}}]{Haibin Zheng} is a PhD student at the college of Information Engineering, Zhejiang University of Technology. He received his bachelor degree from Zhejiang University of Technology in 2017. His research interests include deep learning, artificial intelligence, and adversarial attack and defense.
\end{IEEEbiography}

\vspace{-10 mm}
\begin{IEEEbiography}[{\includegraphics[width=0.95in,height=1.15in,clip,keepaspectratio]{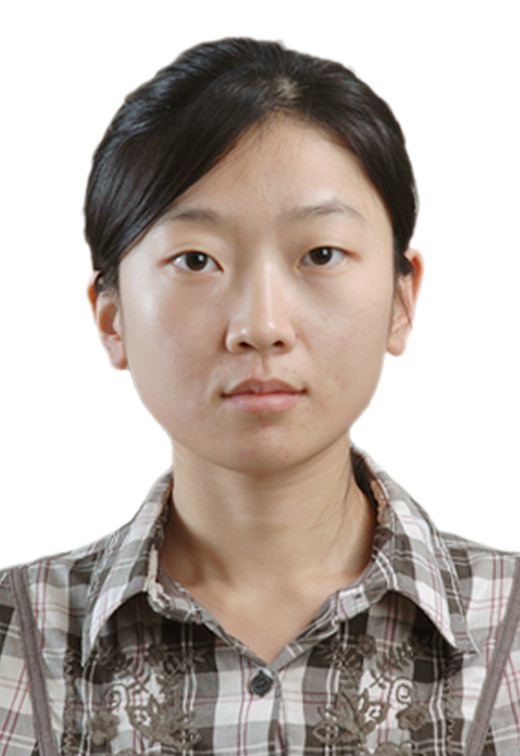}}]{Yao Cheng} is currently a senior researcher at Huawei International in Singapore. She received her Ph.D. degree in Computer Science and Technology from University of Chinese Academy of Sciences. Her research interests include security and privacy in deep learning systems, blockchain technology applications, Android framework vulnerability analysis, mobile application security analysis, and mobile malware detection.
\end{IEEEbiography}







\clearpage
\appendices
\section{Preliminary and Problem Formulation}
\subsection{Vertical Federated Graph Learning}

Denote $G=(V,E)$ is the graph with nodes set $V$ and edges set $E$. Define $A$ as the adjacency of the graph and $X$ as the node features. In VFGL, each client maintains the local GNN with its private data and trains the server model collaboratively. It should be noted that in the existing works~\cite{zhou2020vertically, ni2021vertical}, the edges and node features are split for each client. However, the split subgraph is not in line with the real distribution~\cite{zhang2021federated}. Therefore, we consider the setting of dividing only node features but not edges to preserve the original distribution characteristics of the graph. The server aggregates the local node embeddings uploaded by each client. In this work, we consider concatenating the local embeddings as the aggregation strategy:
\begin{equation}
\begin{split}
            h_{global} &\leftarrow  CONCAT(h_{1},...,h_{i},...,h_{K})\\
    &s.t.\quad h_{i}={f}^{i}_{\theta }(A_{i}, X_{i}, V),
\end{split}
\end{equation}
where $K$ is the number of clients in VFGL, $h_{i}$ is the $i$-th client's local embeddings produced by the local GNN ${f}^{i}_{\theta}(\cdot)$ with the local private data $(A_{i}, X_{i})$.

Then, $h_{global}$ is utilized to train the server model $\mathcal{S}(\cdot)$:
\begin{equation}
    \begin{split}
        \mathcal{L}_{train}&=-\sum_{n=1}^{|V_L|}\sum_{l=1}^{|F|}Y_{nl}\ln(Y^{'}_{nl})\\
        & s.t. \quad Y^{'} = \mathcal{S}(h_{global}),
    \end{split}
\end{equation}
where $V_L$ denotes the subset of $V$ labeled with the ground truth, $|F|$ is the number of node classes. $Y$ is the label list, and $Y^{'}$ is the probabilities list that predicted by the server model. Here, a $L$-layers multilayer perceptron (MLP) is applied as the server model in VFGL in this work.

\subsection{Adversarial Attack on Vertical Federated Graph Learning}
\label{aavfgl}
In the centralized graph learning scenario, the attackers usually reduce the quality of node embeddings to attack the target GNNs by manipulating input data. Likewise, the malicious clients can manipulate the local data and upload the perturbed embeddings to attack the server model with ulterior motives:
\begin{equation}
\begin{split}
    &\hat{Y}_{t}^{'}=\mathcal{S}(\hat{h}_{global})\\
    s.t.\quad &\hat{h}^{v_{t}}_{global}\leftarrow CONCAT(h^{v_{t}}_{1},...,\hat{h}^{v_{t}}_{m},...,h^{v_{t}}_{K})\\
    &\hat{h}^{v_{t}}_{m}={f}^{m}_{\theta}(\hat{A}_{m}, \hat{X}_{m}, v_{t}),
\end{split}
\end{equation}
where ${f}^{m}_{\theta}$ is the local GNN of malicious client,  $\hat{A}_{m}$ and $\hat{X}_{m}$ are perturbed data. $\hat{h}^{v_{t}}_{m}$ is denoted as the low-quality node embeddings of the target node $v_{t}$, which uploaded by the malicious client.

The target of the malicious client is maximizing the loss of $v_{t}$:
\begin{equation}
\begin{split}
    &\max \sum \limits_{v_{t}\in T}{{\mathcal{L}_{atk}}(\mathcal{S}(\hat{h}^{v_{t}}_{global}),{{y}_{v_{t}}})}\\
    &s.t. \quad \mathcal{L}_{atk}=-\sum_{l=1}^{|F|}Y_{v_{t}l}\ln(Y^{'}_{v_{t}l}),
\end{split}
\end{equation}
where $y_{v_{t}}$ is the ground truth of $v_{t}$, and $T$ is the target node set.

\section{Adversarial Attacks}

\begin{itemize}
    \item \textbf{FGA~\cite{chen2018fast}:} It constructs the symmetrical edge gradient matrix of the original graph. Then, the maximal absolute edge gradient guides to generate perturbations iteratively.
    \item \textbf{GradArgmax~\cite{dai2018adversarial}:} It calculates the gradient of edges based on the loss function, and the greedy algorithm is adopted to select the edges to be modified.
    \item \textbf{Nettack~\cite{zugner2018adversarial}:} It selects the candidate edges and node features by graph properties (e.g., degree distribution) and generates adversarial perturbations with the highest scoring edges  iteratively to fool the classifier.
    \item \textbf{SGA~\cite{li2021adversarial}:} It extracts a subgraph centered at the target node first. Then it generates adversarial attacks by flipping edges with the largest gradient in the subgraph.
\end{itemize}

To illustrate the query efficiency of NA$^{2}$, we select a query-based black-box adversarial attack method for comparison.
\begin{itemize}
    \item \textbf{GeneticAlg~\cite{dai2018adversarial}:} It adopts the genetic algorithm to generate the adversarial perturbations, has to query the target model to calculate the fitness score for each candidate solution that is decided by the population.
\end{itemize}

\begin{algorithm}
\label{alo}
\small
\caption{NA$^{2}$}
\KwIn{Original adjacency $A$, original node features $X$, target node $v_t$, starting epoch $\tau$, train set $V_{train}$, attack budget $\Delta$.}
\KwOut{The adversarial adjacency $\hat{A}$.}
Train the VFGL until the starting epoch $\tau$ to start NA$^2$.\\
Divided the train set according to the node categories.\\
\For{$t=1$ to $|V_{train}|$}{Test the train examples according to Equation~\ref{k1st} and~\ref{kother}.\\
Obtain the neuron path $P_t$.}
Count and select the target path $P_T$, and get the candidate nodes set $C$.\\
Identify the significant neuron $k^1$ by $P_T$, and select the top-$M$ features and the target features set.\\
\For{$c$ in $C$}{Modify the features of the candidate nodes following Equation~\ref{mod_fea} and get $\Tilde{X}$.}
Train the VFGL with the manipulated features $\Tilde{X}$ continuously.\\
Query the server model and gain the probabilities $p$.\\
Establish the shadow model $\mathcal{\Tilde{S}}$ with the local GNN and a $L$-layers MLP.\\
\For{t=1 to $T_{shadow}$}{
Train $\mathcal{\Tilde{S}}$ to minimize the MSE in Equation~\ref{loss_shadow}.
}
Input the original adjacency, the manipulated features $\Tilde{X}$ and the shadow $\mathcal{\Tilde{S}}$ into the centralized attack method.\\
\For{i=1 to $\Delta$}{Generate the adversarial adjacency $\hat{A}$ as Equation~\ref{ad_emb}.}
\textbf{return} the adversarial adjacency matrix $\hat{A}$.
\end{algorithm}

\end{document}